\documentclass{article}

\usepackage[table]{xcolor}
\usepackage{graphicx}
\usepackage{microtype}
\usepackage{subfigure}
\usepackage{booktabs} %
\usepackage{hyperref}
\usepackage{url}
\usepackage{tabularx}
\usepackage{adjustbox}
\usepackage{multirow}
\usepackage{makecell}
\usepackage{amsmath}
\usepackage{caption} 
\usepackage{xcolor}
\usepackage{wrapfig,lipsum}
\usepackage[table]{xcolor}

\definecolor{darkgreen}{rgb}{0.0, 0.8, 0.0}
\usepackage{hyperref}
\usepackage{float}

\newcommand{\revise}[1]{\textcolor{black}{#1}}

\newcommand{\keypoint}[1]{\noindent\textbf{#1}\quad} 

\usepackage[accepted]{icml2024}

\usepackage{amsmath}
\usepackage{amssymb}
\usepackage{mathtools}
\usepackage{amsthm}

\usepackage[capitalize,noabbrev]{cleveref}

\theoremstyle{plain}

\theoremstyle{definition}

\theoremstyle{remark}

\usepackage[textsize=tiny]{todonotes}

\icmltitlerunning{Fool Your (Vision and) Language Model with Embarrassingly Simple Permutations}

\begin{document}

\twocolumn[
\icmltitle{Fool Your (Vision and) Language Model with \\Embarrassingly Simple Permutations}

\icmlsetsymbol{equal}{*}

\begin{icmlauthorlist}
\icmlauthor{Yongshuo Zong}{ed}
\icmlauthor{Tingyang Yu}{epfl}
\icmlauthor{Ruchika Chavhan}{ed}
\icmlauthor{Bingchen Zhao}{ed}
\icmlauthor{Timothy Hospedales}{ed}
\end{icmlauthorlist}

\icmlaffiliation{ed}{University of Einburgh}
\icmlaffiliation{epfl}{EPFL}

\icmlcorrespondingauthor{Yongshuo Zong}{yongshuo.zong@ed.ac.uk}

\icmlkeywords{Machine Learning, ICML}

\vskip 0.3in
]

\printAffiliationsAndNotice{}  %

\begin{abstract}
Large language and vision-language models are rapidly being deployed in practice thanks to their impressive capabilities in instruction following, in-context learning, and so on. This raises an urgent need to carefully analyse their robustness so that stakeholders can understand if and when such models are trustworthy enough to be relied upon in any given application. 
In this paper, we highlight a specific vulnerability in popular models, namely permutation sensitivity in multiple-choice question answering (MCQA). 
Specifically, we show empirically that popular models are vulnerable to adversarial permutation in answer sets for multiple-choice prompting, which is surprising as models should ideally be as invariant to prompt permutation as humans are.  
These vulnerabilities persist across various model sizes, and exist in very recent language and vision-language models. Code is available at \url{https://github.com/ys-zong/FoolyourVLLMs}. %

\end{abstract}

\section{Introduction}
Large language models (LLMs)~\citep{brown2020language, openai2023chatgpt, touvron2023llama} and large vision-language models (VLLMs)~\citep{alayrac2022flamingo, li2023blip} have made astonishing progress in recent years. They have attained strong capabilities across a diverse array of language tasks, enabling nuanced text generation, sophisticated instruction following, and natural dialogue with multimodal input and output. One task where they demonstrate particular prowess is multiple-choice question answering (MCQA)~\citep{robinson2023leveraging}. This is an important capability with many real-world applications, from education to recruitment exams. Current LLMs and VLLMs have widely utilized the task format of MCQA for benchmarking and evaluation~\citep{hendrycks2020measuring, lu2022learn, zhong2023agieval, liang2022holistic, schwenk2022okvqa}. This has built confidence that they can generate accurate and robust answers, underpinned claims of LLM competence at professional level human qualifications such as the bar exam \citep{openai2023gpt4}, and even led to reports of surpassing human-level performance on various tasks. 

\begin{figure}
    \centering
    \includegraphics[width=\linewidth]{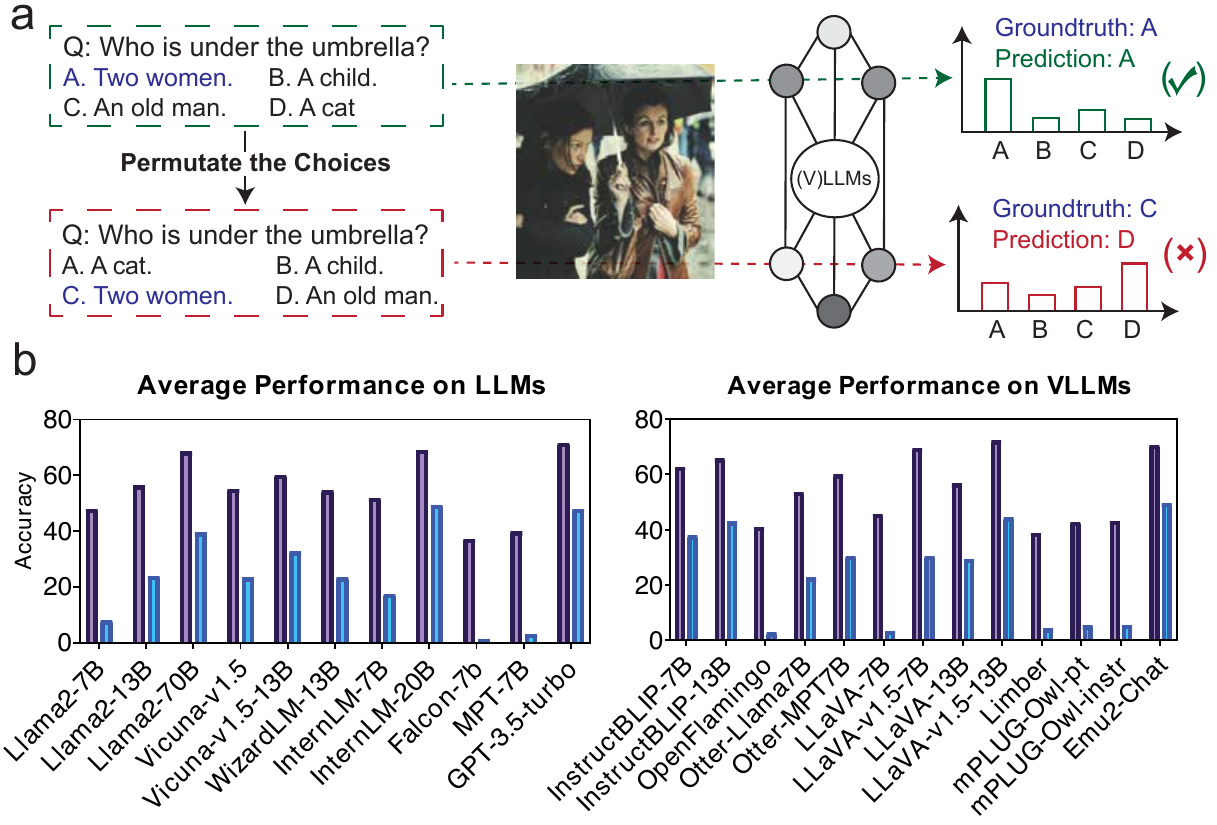}
    \caption{\textbf{a.} Schematic Illustration of an MCQA permutation attack. \textbf{b.} Summary of MCQA adversarial attack results for both LLMs and VLLMs. The values are average accuracy across all benchmarking datasets.}
    \label{fig:teaser}
\end{figure}

Surprisingly, contrary to the confidence instilled by high-performance metrics on established benchmarks, these models are surprisingly brittle when subjected to simple permutations in the answer choices, i.e., randomly changing the option positions. In this paper, we show that even a simple permutation of the answer sets, as illustrated in Figure~\ref{fig:teaser}, can lead to a dramatic decline in accuracy for both LLMs and VLLMs in a wide range of MCQA datasets, sometimes even below the random chance levels.  
For instance, Llama2-13B~\citep{touvron2023llama} experiences a 33.89\% degradation in accuracy on the MMLU dataset~\citep{hendrycks2020measuring} following random permutation of option positions, with results falling below the random chance. A wide variety of popular LLMs and VLLMs, suffer significantly from this vulnerability, as summarised in Figure~\ref{fig:teaser} b.

Furthermore, our investigations reveal an even more disconcerting aspect: the vulnerability to permutations persists in LLMs and VLLMs even when multiple distractor options are deliberately removed from the answer sets. Intuitively, one expects that by eliminating incorrect choices, the task should become simpler due to increasing chance performance, thereby enhancing the models' performance. However, our empirical findings contradict this notion. Even with a reduced number of distractors, the performance of both LLMs and VLLMs remains susceptible to degradation, affirming the deeply ingrained nature of this vulnerability.

To further investigate the source of the brittleness, we demonstrate through our adversarial attack that it is not merely a selection bias towards/against certain positions, such as moving correct answers to a fixed position that a given model is biased against picking. While positional factors may moderately influence model performance, they do not explain the strength of our adversarial attack results, suggesting a more systemic issue that extends beyond simple position bias. 

This issue should be of intrinsic concern to those seeking to understand and design trustworthy and reliable LLMs and VLLMs, or emulate human capabilities. However, one might speculate that the issue could be mitigated in practice through the engineering solution of majority voting across different permutations or by employing calibration strategies as suggested in previous work~\citep{zhao2021calibrate}. However, our findings indicate that while majority voting may offer some degree of improvement, the resulting performance still lags behind the original metrics, despite incurring a $k!\times$ computational cost of the original inference time. Additionally, calibration techniques such as calibrate-before-use \citep{zhao2021calibrate} fail to alleviate this problem effectively.

In summary, our research unveils a glaring yet often overlooked vulnerability in large language models and vision-language models, specifically within the domain of multiple-choice question answering (MCQA). Despite their impressive metrics on well-established benchmarks, these models reveal a disconcerting fragility when faced with simple manipulations such as option permutations. Existing mitigation strategies fall short of effectively resolving this issue. Our observations not only raise pivotal questions about the models' robustness but also accentuate the necessity for heightened scrutiny in assessing their MCQA capabilities. We argue that stakeholders should be vigilant in relying on such models until these vulnerabilities are adequately addressed.

\section{Simple Adversarial Attack Breaks LLMs and VLLMs}
In this section, we analyse the brittleness of a broad array of large language models and vision-language models to random adversarial attacks in MCQA. By simply shuffling answer choices, we find that these models fail to maintain their performance, revealing a critical vulnerability.

\subsection{Experiment Setup}
In an ideal scenario, robust models should offer consistent predictions that are \emph{invariant} to permutations that have no semantic influence on the question being posed. To test this, we simply iterate through the possible permutations of MCQ options. A robust model should be correct in every case. While there are $k!$ possible combinations in total, we cease permutation once the model produces an incorrect prediction (succumbs to the permutation attack), which usually requires far less than $k!$ attempts\footnote{Since typical MCQA benchmarks use $k=4$, the brute force algorithm is cheaper than a gradient-based solution. But gradient-based solutions could be used if the attack needs to scale to substantially larger $k$.}. 

Formally, Given a question \( q \) and an answer list \( A = \{a_1, a_2, \ldots, a_k\} \), the permutation adversarial attack can be described by the Equation~\ref{eq:attack}. We maximize the loss function ($\mathcal{L}$) with respect to all possible permutations  ($\Pi$) of the answer list. Here, \( \text{prompt}(q, A) \) prompts the model with the given query and answer list, and the model's response is then evaluated by the loss. 
\begin{align}
\label{eq:attack}
\text{Maximize:} \quad \mathcal{L}\left( \text{prompt}(q, A^* \right)) \nonumber \\
\text{s.t.} \quad A^* \in \Pi(A)
\end{align}

\begin{table}[h]
\caption{Statistics of the language datasets evaluated.}
\centering
\small
\setlength{\tabcolsep}{2pt}
\renewcommand{\arraystretch}{1} %
\begin{tabular}{cccc}
\toprule
 & {\# Choices} & {\# QA Pairs} & {Task} \\
\midrule
\bf MMLU & 4 & 14079 & {Aggregated} \\
\bf ARC-c & 4 & 1165 & {Commonsense Reasoning} \\
\bf BoolQ & 2 & 3270 & {Reading Comprehension} \\
\bf SocialiQA & 3 & 1954 & {Commonsense Reasoning} \\
\bf MedMCQ & 4 & 2816 & {Out-of-domain} \\
\bottomrule
\end{tabular}
\label{table:l-dataset}
\end{table}

\begin{table}[h]
\setlength{\tabcolsep}{2pt}
\caption{Statistics of the vision-language datasets evaluated.}
\begin{center}
\resizebox{\linewidth}{!}{%
\begin{tabular}{lccc}
\toprule
{\bf} &  \# Choices &  \# QA pairs & Task \\
\midrule
{\bf ScienceQA} & 2,3,4,5 & 2021 & Scientific QA\\
{\bf A-OKVQA} & 4 & 1145 & Commonsense Reasoning \\
{\bf MMBench} & 4 & 4377 & Aggregated \\
{\bf SEED-Bench} & 4 & 14233 & Aggregated \\
\bottomrule
\end{tabular}
}
\end{center}
\label{table:vl-dataset}
\end{table}

\keypoint{Models} We evaluate a wide range of LLMs and VLLMs of diverse sizes, different pretrained backbones, and both auto-regressive pretrained and instruction-following fine-tuned models. Specifically, for LLMs, we have evaluated LLaMA-2 (7B/13B)~\citep{touvron2023llama2}, Vicuna (7B/13B)~\citep{chiang2023vicuna}, WizardLM-13B~\citep{xu2023wizardlm}, InternLM-20B~\citep{2023internlm}, Falcon-7B~\cite{penedo2023refinedwebfalcon}, and MPT-7B~\citep{MosaicML2023Introducing}. For VLLMs, InstructBLIP (Vicuna-based, 7B/13B)~\citep{dai2023instructblip}, Open-Flamingo (MPT-based, 9B)~\citep{awadalla2023openflamingo}, Otter (Llama-based, MPT-based)~\citep{li2023otter}, LLaVA (7B/13B)~\citep{liu2023llava}, LLaVA-v1.5 (7B/13B)~\citep{liu2023improved}, Emu2-Chat~\citep{sun2024emu2}, Limber (7B)~\citep{merullo2022linearly}, and mPLUG-Owl (pretraining, intruction)~\citep{ye2023mPLUGowl} are used for evaluation.

\keypoint{Datasets} We utilize a diverse array of language and vision-language MCQA datasets for comprehensive evaluation. These datasets cover multiple domains and require different aspects of the models to give correct answers, ensuring our findings are generalizable. Specifically, for LLMs, we utilize MMLU~\citep{hendrycks2020measuring}, ARC challenge (ARC-c)~\citep{clark2018arc}, BoolQ~\citep{clark2019boolq}, SocialiQA~\citep{sapetal2019social}, and MedMCQA~\citep{pal2022medmcqa}. For VLLMs, we use ScienceQA~\citep{lu2022learn}, A-OKVQA~\citep{schwenk2022okvqa}, MMBench~\citep{liu2023mmbench}, and SEED-Bench~\citep{li2023seed}. We use the questions in ScienceQA that have corresponding images, the MCQA subsets of MMBench, and the image-based MCQAs in SEED-Bench.

\keypoint{Evaluations}
We use accuracy as our primary metric. During testing, we prompt the model to generate the possible option symbols (e.g., A to D) and extract the probability assigned to each choice in the first position. The option with the highest probability is then selected as the model's answer for that specific question. For both LLMs and VLLMs, we use greedy decoding to ensure reproducibility. All of the datasets we use are publicly available. All of the model weights (except GPT-3.5-Turbo) can be obtained from the HuggingFace model zoo or the original official Github repositories. GPT-3.5-Turbo can be accessed from OpenAI API. Experiments are conducted on A100-80GB GPUs.

\subsection{Main Results}
We present the main results in Table~\ref{table:LLMs} and~\ref{table:VLLMs} for language and vision-language models respectively. 

\textbf{Language Models}\quad In our experiments, large language models manifested a significant susceptibility to adversarial permutations, a finding consistent across various MCQA benchmarks. Our evaluation extended beyond the typical four-option MCQA datasets to include more diverse formats like the two-option BoolQ~\citep{clark2019boolq} and the three-option SocialIQA~\citep{sapetal2019social} that are naturally more resilient to the permutations. Intriguingly, the presence of only one or two distractor options did not mitigate the model's vulnerability to permutations. For instance, Llama2-7B's accuracy on BoolQ plummeted from 61.79\% to a mere 8.23\%, a performance even worse than random chance. Moreover, out of 50 experiments conducted with large language models, only 12 non-GPT-3.5-turbo models managed to perform better than random chance. And all of them, including GPT-3.5-turbo, suffer from significant performance decreases.

\textbf{Vision-Language Models}\quad
In the vision-language model evaluations, the susceptibility to adversarial permutations is also severe. Despite the presence of visual context, which may intuitively add a layer of resilience, the VLLMs were not spared from the adverse effects of our permutation attacks. Among 48 experiments, more than half of them fell below random chance performance after the adversarial attack. While InstructBLIP~\citep{dai2023instructblip} and LLaVA-v1.5~\citep{liu2023improved} show relatively strong robustness to the adversarial attack, all of the models experienced significant accuracy drops ranging from 20\% to 45\%.

\textbf{Further Observations}\quad
We note that within the same model family but with varying parameter sizes (e.g., InstructBLIP-7B v.s. InstructBLIP-13B), scaling up generally enhances both the baseline performance and resilience to adversarial attacks with relatively smaller declines in accuracy. We can also observe that models have different success rates over random chance in different datasets. For example, all of the LLMs failed the adversarial attack on MedMCQA dataset except GPT-3.5-turbo, which is also only slightly above the random chance. It shows the challenges of LLMs to generalize to out-of-domain data, and suggests caution about their use in unconstrained practical scenarios. %

\begin{table*}[t]
\caption{Performance comparisons of LLMs before and after adversarial attack. Numbers in each represent original accuracy, accuracy after adversarial attack, and relative performance drop. Red shading indicates experiments where the permutation attack reduced performance below chance level. All models suffer substantially with most experiments leading to below chance performance.}
\begin{center}
\resizebox{\linewidth}{!}{%
\begin{tabular}{llllllll}
\toprule
{\bf Method} & {\bf MMLU} & {\bf ARC-c} & {\bf BoolQ}  &  {\bf SocialiQA }& {\bf MedMCQA}\\
\midrule

Llama2-7B & \cellcolor{red!10}40.91/\phantom{0}6.17 (\textcolor{red}{34.74 $\downarrow$}) & \cellcolor{red!10}47.04/\phantom{0}7.98 (\textcolor{red}{39.06 $\downarrow$}) & \cellcolor{red!10}61.79/\phantom{0}8.23 (\textcolor{red}{53.56 $\downarrow$}) &\cellcolor{red!10}52.00/15.71 (\textcolor{red}{36.29 $\downarrow$}) &\cellcolor{red!10}37.96/\phantom{0}1.60 (\textcolor{red}{36.36 $\downarrow$}) \\[4pt]

Llama2-13B & \cellcolor{red!10}52.22/18.33 (\textcolor{red}{33.89 $\downarrow$}) & \cellcolor{red!10}61.80/21.63 (\textcolor{red}{40.17 $\downarrow$}) & \cellcolor{red!10}67.16/38.29 (\textcolor{red}{28.87 $\downarrow$}) & 61.21/34.14 (\textcolor{red}{27.07 $\downarrow$}) & \cellcolor{red!10}39.78/\phantom{0}7.35 (\textcolor{red}{32.43 $\downarrow$}) \\[4pt]

\revise{Llama2-70B} & 64.68/33.16 (\textcolor{red}{31.52 $\downarrow$}) & 80.00/51.50 (\textcolor{red}{28.50 $\downarrow$}) & 76.39/56.21 (\textcolor{red}{20.18 $\downarrow$}) & 71.60/49.85 (\textcolor{red}{21.75 $\downarrow$}) & \cellcolor{red!10}49.61/\phantom{0}7.35 (\textcolor{red}{32.43 $\downarrow$}) \\[4pt]

Vicuna-v1.5 & \cellcolor{red!10}48.57/18.09 (\textcolor{red}{30.48 $\downarrow$}) & \cellcolor{red!10}58.37/23.43 (\textcolor{red}{34.94 $\downarrow$}) & \cellcolor{red!10}64.04/29.60 (\textcolor{red}{34.44 $\downarrow$}) & 64.99/38.33 (\textcolor{red}{26.66 $\downarrow$}) & \cellcolor{red!10}39.28/\phantom{0}7.67 (\textcolor{red}{31.61 $\downarrow$}) \\[4pt]

Vicuna-v1.5-13B & 54.68/26.27 (\textcolor{red}{28.41 $\downarrow$}) & 69.27/38.80 (\textcolor{red}{30.47 $\downarrow$}) &  \cellcolor{red!10}68.96/42.14 (\textcolor{red}{26.82 $\downarrow$}) & 66.07/44.42 (\textcolor{red}{21.65 $\downarrow$}) & \cellcolor{red!10}41.80/11.90 (\textcolor{red}{29.90 $\downarrow$}) \\[4pt]

WizardLM-13B & \cellcolor{red!10}48.60/15.87 (\textcolor{red}{32.73 $\downarrow$}) & \cellcolor{red!10}58.20/21.12 (\textcolor{red}{37.08 $\downarrow$}) & \cellcolor{red!10}67.49/42.11 (\textcolor{red}{25.38 $\downarrow$}) & \cellcolor{red!10}63.46/31.78 (\textcolor{red}{31.68 $\downarrow$}) & \cellcolor{red!10}34.87/\phantom{1}6.32 (\textcolor{red}{28.55 $\downarrow$}) \\[4pt]

\revise{InternLM-7B} & \cellcolor{red!10}45.72/10.45 (\textcolor{red}{35.27 $\downarrow$}) & \cellcolor{red!10}56.14/17.34 (\textcolor{red}{38.80 $\downarrow$}) & \cellcolor{red!10}65.83/26.41 (\textcolor{red}{39.42 $\downarrow$}) & \cellcolor{red!10}59.47/30.30 (\textcolor{red}{29.17 $\downarrow$}) & \cellcolor{red!10}32.63/\phantom{1}2.56 (\textcolor{red}{30.07 $\downarrow$}) \\[4pt]

InternLM-20B & 59.14/29.52 (\textcolor{red}{29.62 $\downarrow$}) & 78.28/54.42 (\textcolor{red}{23.86 $\downarrow$}) & 85.20/82.91 (\textcolor{red}{\phantom{0}2.29 $\downarrow$}) & 79.48/65.97 (\textcolor{red}{13.51 $\downarrow$}) & \cellcolor{red!10}43.61/13.92 (\textcolor{red}{29.69 $\downarrow$}) \\[4pt]

Falcon-7b & \cellcolor{red!10}31.66/\phantom{0}2.49 (\textcolor{red}{29.17 $\downarrow$}) & \cellcolor{red!10}34.74/\phantom{0}0.09 (\textcolor{red}{34.65 $\downarrow$}) \cellcolor{red!10}&\cellcolor{red!10} 55.35/\phantom{0}2.66 (\textcolor{red}{52.69 $\downarrow$}) & \cellcolor{red!10}36.29/\phantom{0}0.55 (\textcolor{red}{35.74 $\downarrow$}) & \cellcolor{red!10}28.12/\phantom{0}0.07 (\textcolor{red}{28.05 $\downarrow$}) \\[4pt]

MPT-7B & \cellcolor{red!10}35.60/\phantom{0}3.52 (\textcolor{red}{32.08 $\downarrow$}) & \cellcolor{red!10}37.76/\phantom{0}1.06 (\textcolor{red}{36.70 $\downarrow$}) &\cellcolor{red!10}58.46/\phantom{0}7.03 (\textcolor{red}{51.43 $\downarrow$}) & \cellcolor{red!10}41.61/\phantom{0}2.53 (\textcolor{red}{39.08 $\downarrow$}) & \cellcolor{red!10}26.31/\phantom{0}1.60 (\textcolor{red}{24.71 $\downarrow$}) \\[4pt]

GPT-3.5-turbo & 64.81/40.39 (\textcolor{red}{24.42 $\downarrow$}) & 82.23/61.55 (\textcolor{red}{20.68 $\downarrow$}) & 87.92/81.35 (\textcolor{red}{\phantom{0}6.57 $\downarrow$}) & 70.62/56.29 (\textcolor{red}{14.33 $\downarrow$}) & 52.22/32.07 (\textcolor{red}{20.15 $\downarrow$}) \\[4pt]
Random Chance & 25.0  & 25.0  & 50.0 &  33.33 & 25.0 \\

\bottomrule
\end{tabular}
}
\end{center}
\label{table:LLMs}
\end{table*}

\begin{table*}[h]
\caption{Performance comparisons of VLLMs before and after adversarial attack. Numbers in each cell represent original accuracy, accuracy after adversarial attack, and relative performance drop. Red shading indicates performance below chance level after the permutation attack. All models suffer substantially with most experiments leading to below chance performance.}
\begin{center}
\resizebox{0.95\linewidth}{!}{%
\begin{tabular}{lllllll}
\toprule
{\bf Method} & {\bf ScienceQA} & {\bf A-OKVQA} & {\bf SEED-Bench} & {\bf MMBench}   \\ %
\midrule
   
 InstructBLIP-7B & 59.46/33.31 (\textcolor{red}{26.15 $\downarrow$}) & 74.06/51.62 (\textcolor{red}{22.44 $\downarrow$}) & 51.61/25.68 (\textcolor{red}{25.93 $\downarrow$}) & 64.91/41.01 (\textcolor{red}{23.90 $\downarrow$}) \\[4pt]

 InstructBLIP-13B  & 64.15/41.84 (\textcolor{red}{22.31 $\downarrow$}) & 77.90/55.38 (\textcolor{red}{22.52 $\downarrow$}) & 53.65/28.79 (\textcolor{red}{24.86 $\downarrow$}) & 67.12/45.49 (\textcolor{red}{21.63 $\downarrow$}) \\[4pt]
 
 OpenFlamingo & \cellcolor{red!10}39.43/1.37 (\textcolor{red}{38.06 $\downarrow$}) & \cellcolor{red!10}46.90/3.58 (\textcolor{red}{43.32 $\downarrow$}) &\cellcolor{red!10} 37.99/0.87 (\textcolor{red}{37.12 $\downarrow$})  & \cellcolor{red!10}38.99/5.18 (\textcolor{red}{33.81 $\downarrow$})\\[4pt]
  
 Otter-Llama7B & 59.92/32.54 (\textcolor{red}{27.38 $\downarrow$}) & 57.99/28.30 (\textcolor{red}{29.69 $\downarrow$}) & \cellcolor{red!10}40.77/9.91 (\textcolor{red}{30.86 $\downarrow$}) & \cellcolor{red!10}55.24/19.67 (\textcolor{red}{35.57 $\downarrow$}) \\[4pt]

  Otter-MPT7B  & 63.11/31.38 (\textcolor{red}{31.73 $\downarrow$}) & 68.21/43.19 (\textcolor{red}{25.02 $\downarrow$}) & \cellcolor{red!10}46.76/10.82 (\textcolor{red}{35.94 $\downarrow$})& 61.31/36.46 (\textcolor{red}{24.85 $\downarrow$}) &\\[4pt]
  
  LLaVA-7B & \cellcolor{red!10}45.20/2.28 (\textcolor{red}{42.92 $\downarrow$}) & \cellcolor{red!10}52.91/\phantom{0}0.09 (\textcolor{red}{52.82 $\downarrow$}) & \cellcolor{red!10}38.36/5.67 (\textcolor{red}{43.03 $\downarrow$}) & \cellcolor{red!10}46.03/5.07 (\textcolor{red}{40.96 $\downarrow$}) \\[4pt]

  LLaVA-13B & 60.63/46.53 (\textcolor{red}{14.10 $\downarrow$}) & 63.14/25.85 (\textcolor{red}{37.29 $\downarrow$}) &\cellcolor{red!10}44.00/13.68 (\textcolor{red}{30.32 $\downarrow$}) & 59.13/31.30 (\textcolor{red}{27.83 $\downarrow$})\\[4pt]

LLaVA-v1.5-7B &  67.78/45.61 (\textcolor{red}{22.17 $\downarrow$}) & 81.31/36.24 (\textcolor{red}{45.07 $\downarrow$}) & \cellcolor{red!10}59.17/12.44 (\textcolor{red}{46.73 $\downarrow$}) & 69.57/27.39 (\textcolor{red}{42.18 $\downarrow$})\\[4pt]

LLaVA-v1.5-13B & 71.60/52.55 (\textcolor{red}{19.05 $\downarrow$})& 83.32/47.25 (\textcolor{red}{36.07 $\downarrow$})& \cellcolor{red!10}61.50/18.95 (\textcolor{red}{42.55 $\downarrow$})& 72.33/58.42 (\textcolor{red}{13.91 $\downarrow$})\\[4pt]

 Limber & \cellcolor{red!10}49.33/14.03 (\textcolor{red}{35.30 $\downarrow$}) & \cellcolor{red!10}39.57/1.22 (\textcolor{red}{38.35 $\downarrow$}) & \cellcolor{red!10}31.50/0.26 (\textcolor{red}{31.24 $\downarrow$})  & \cellcolor{red!10}34.93/1.62 (\textcolor{red}{33.31 $\downarrow$})  \\[4pt]

 mPLUG-Owl-pt & \cellcolor{red!10}53.24/10.20 (\textcolor{red}{43.04 $\downarrow$}) &\cellcolor{red!10}39.91/1.83 (\textcolor{red}{38.08 $\downarrow$}) & \cellcolor{red!10}35.57/0.91 (\textcolor{red}{34.66 $\downarrow$}) & \cellcolor{red!10}42.57/8.54 (\textcolor{red}{34.03 $\downarrow$})\\[4pt]

 mPLUG-Owl-instr & \cellcolor{red!10}54.87/11.43 (\textcolor{red}{43.44 $\downarrow$}) &  \cellcolor{red!10}37.12/2.01 (\textcolor{red}{35.11 $\downarrow$}) & \cellcolor{red!10}36.74/2.72 (\textcolor{red}{34.02 $\downarrow$}) & \cellcolor{red!10}43.74/6.12 (\textcolor{red}{37.62 $\downarrow$}) \\[4pt]

 Emu2-Chat & 64.60/44.27 (\textcolor{red}{20.33$\downarrow$}) & 81.91/63.67 (\textcolor{red}{ 18.24$\downarrow$}) & 62.11/38.02 (\textcolor{red}{24.09$\downarrow$}) & 73.21/52.44 (\textcolor{red}{20.77$\downarrow$}) \\[4pt] 

 Random Chance & Min 20.0 & 25.0 & 25.0 & 25.0 \\

\bottomrule
\end{tabular}
}
\end{center}
\label{table:VLLMs}
\end{table*}

\subsection{Answer Set Pruning}
\label{ans_prune}

In this subsection, we examine the impact of a stricter test condition on MCQA, specifically by reducing the number of distractor options, while obviously retaining the true answer. This is expected to improve baseline performance by increasing random chance level, but also we expected it to reduce vulnerability to adversarial permutation by substantially reducing the degrees of freedom that the permutation attack can explore. However, we found that models remain highly susceptible to even the few permutations available in the reduced set of options.

\textbf{Experiment Setup}\quad 
Specifically, we constrain the answer set by reducing the number of total choices from four to either three or two, inclusive of the ground-truth answer. We then compare the performance metrics between these pruned sets in both permuted and non-permuted conditions to assess the relative susceptibility of the models. 

\textbf{Results}\quad 
We present the results of answer set pruning of MMLU datasets in Table~\ref{table:LLMs-prune} and other datasets in the appendix. As can be seen from Table~\ref{table:LLMs-prune}, reducing the number of options increases the base prediction accuracy as expected, but performing adversarial permutation on the reduced answer set still dramatically reduces the accuracy even in the 2-option cases. In most cases, the performance is below the chance level given the number of options. This means that, surprisingly, even in the simplest case of a binary choice, models are not robust to whether the true answer is presented as the first or second option.

\begin{table}[ht]
\caption{Performance of LLMs on the MMLU dataset under answer set pruning. Numbers in each cell represent original accuracy, accuracy after adversarial attack, and relative performance drop. Baseline performances improve as the number of distractors is reduced, but performance is reduced below chance after adversarial permutation.}
\begin{center}
\resizebox{\linewidth}{!}{%
\begin{tabular}{llll}
\toprule
{\bf Method} & {\bf 4 Choices} & {\bf 3 Choices} & {\bf 2 Choices}\\
\midrule

Llama2-7B & 40.91 & \cellcolor{red!10}48.75/\phantom{0}8.67 (\textcolor{red}{{39.08$\downarrow$}}) & \cellcolor{red!10}63.33/20.26 (\textcolor{red}{{43.07$\downarrow$}}) \\ [4pt]
Llama2-13B & 52.22 & \cellcolor{red!10}70.77/22.85 (\textcolor{red}{{47.92$\downarrow$}}) & \cellcolor{red!10}71.13/31.85 (\textcolor{red}{{39.28$\downarrow$}}) \\ [4pt]

\revise{Llama2-70B} & 64.68 & 69.90/35.34 (\textcolor{red}{{34.56$\downarrow$}}) &  \cellcolor{red!10}75.23/45.88 (\textcolor{red}{{29.35$\downarrow$}})  \\[4pt]

Vicuna-v1.5-7B & 48.57 &\cellcolor{red!10}56.65/30.60 (\textcolor{red}{{26.97$\downarrow$}}) & \cellcolor{red!10}68.81/32.60 (\textcolor{red}{{36.21$\downarrow$}}) \\[4pt]
Vicuna-v1.5-13B & 54.68 & \cellcolor{red!10}61.75/29.02 (\textcolor{red}{{32.66$\downarrow$}}) & \cellcolor{red!10}72.97/28.06 (\textcolor{red}{{44.91$\downarrow$}}) \\[4pt]
WizardLM-13B & 48.60 & \cellcolor{red!10}56.57/17.74 (\textcolor{red}{{38.83$\downarrow$}}) & \cellcolor{red!10}69.09/28.96 (\textcolor{red}{{40.13$\downarrow$}}) \\[4pt]
\revise{InternLM-7B} & 45.72 & \cellcolor{red!10}51.76/12.39 (\textcolor{red}{39.37$\downarrow$}) & \cellcolor{red!10}65.88/19.65 (\textcolor{red}{46.23$\downarrow$}) \\[4pt]
InternLM-20B & 59.14 & \cellcolor{red!10}65.25/30.48 (\textcolor{red}{{34.67$\downarrow$}}) & \cellcolor{red!10}76.09/43.51 (\textcolor{red}{{32.58$\downarrow$}}) \\[4pt]
Falcon-7b & 31.66 &\cellcolor{red!10}52.88/\phantom{0}5.92 (\textcolor{red}{{46.96$\downarrow$}}) & \cellcolor{red!10}58.31/11.41 (\textcolor{red}{{46.90$\downarrow$}}) \\[4pt]
MPT-7B &  35.60& \cellcolor{red!10}53.31/\phantom{0}6.27 (\textcolor{red}{{47.03$\downarrow$}}) & \cellcolor{red!10}58.31/15.44 (\textcolor{red}{{42.87$\downarrow$}}) \\[4pt]
\revise{GPT-3.5-turbo} & 64.81 & 70.80/42.99 (\textcolor{red}{{27.81$\downarrow$}}) &  79.30/50.82 (\textcolor{red}{{28.48$\downarrow$}})\\[4pt]
Random Chance & 25.0 & 33.33 & 50.0 \\
\bottomrule
\end{tabular}
}
\end{center}
\label{table:LLMs-prune}
\end{table}

\section{Understanding Vulnerability Causes}
\label{deeper}
In this section, we delve into a detailed analysis of the potential causes behind the demonstrated vulnerability and examine related attack types and potential symbol-content spurious correlation. We also refer readers to the Appendix~\ref{app:prompting} for the effect of different prompting and attack techniques and~\ref{app:qualitative}for qualitative results.

\subsection{Position Bias and Other Attacks}
A concurrent study to ours argued for the existence of \emph{position bias} in language model MCQA \citep{zheng2023sensitiveMCQ}. For example, in an A/B/C/D MCQ situation, a given model might have a predisposition to selecting a particular option such as ``C" and an aversion to selecting some other option such as ``A", irrespective of the correctness of the answer associated with each label. Position bias could potentially explain adversarial permutation vulnerability if a model is so averse to selecting a particular option, that rotating the true answer into that slot would reliably cause it to fail. 

To analyse whether position bias can explain our results, we compare our adversarial permutation results to the performance of each LLM under position bias analysis -- always rotating the correct answer to a specific slot (A/B/C/D) in the answer list. 

From the results in Table~\ref{table:LLMs-pos}, we do see the position bias effect remarked upon by \citet{zheng2023sensitiveMCQ}. The models tested exhibit varying degrees of position bias, as results fluctuate with respect to original performance (left column). For example, Vicuna suffers limited position bias, while Falcon-7B is highly position biased. Falcon-7B's baseline accuracy of 31\% rises to  70.9\% when the true answer is placed in slot A -- indicating a strong preference for choosing A; but drops to 3.7\% when the true answer is placed in slot B, indicating a strong aversion to selecting B. 

Comparing the observed position bias to the impact of our adversarial permutation, we can see that our adversarial permutation has a much stronger effect. The results after permutation (right column) are substantially worse than the position bias results. For example, Llama2-7B performs above chance level for answers in every possible position (A/B/C/D), but is reduced to below chance by our adversarial permutation. Thus we conclude that \emph{the impact of our adversarial permutation is not explainable by position bias}. Evidently, models rely on the relationships between choices, including the distractors, which the adversarial permutation manipulates to fool them. I.e., it is not just the true answer, and the location of the true answer (position bias), but also the pattern of the distractor answers around the true answer (as explored by adversarial permutations) that determine model success or failure. This reveals a complex and concerning form of vulnerability. 

\revise{Additionally, to further investigate the potential causes of the vulnerability and compare with other types of attacks, we consider circular evaluation (CircularEval)~\citep{liu2023mmbench} and symbol attack. Specifically, CircularEval involves rotating options while maintaining their relative positions. Symbol attack refers to using different option symbols (here we consider A/B/C/D, a/b/c/d, I/II/III/IV). In both cases, the predictions are counted as correct only if the model predicts all of the variations correctly. As shown in Table~\ref{table:LLMs-pos}, while these attacks degrade performance to some extent, our adversarial attack exhibits the most substantial impact and causes the largest performance drop.
}

\begin{table*}[t]
\caption{\revise{Comparison of positional bias, circular evaluation, symbol attack, and our adversarial permutation on MMLU dataset. Position bias and other attacks have moderate impact.} In contrast, our adversarial permutation severely degrades performance, usually below random chance level.}
\begin{center}
\resizebox{0.9\linewidth}{!}{%
\begin{tabular}{llllllccc}
\toprule
{\bf Method}& {\bf Original} & {\bf A} & {\bf B} & {\bf C} & {\bf D} & {\bf \revise{CircularEval}} & {\bf \revise{Symbol Attack}} & {\bf Permutation Attack}\\
\midrule
  Llama2-7B   &  40.91 &  60.02 & 37.28 & 30.69 & 35.43 & 27.26 & 25.70 & \phantom{0}\textbf{6.17} \\[4pt]

  Llama2-13B  &  52.22 & 36.15  & 58.69 & 59.08 & 54.91 & 35.80 & 30.76 & \textbf{18.33} \\[4pt]

  \revise{Llama2-70B}  &  64.68 &  63.63  &   64.28 &  67.45 & 62.43  & 48.18 & 47.40 & \textbf{33.16} \\[4pt]

 Vicuna-7B    &  48.57 &  49.83 & 63.22 & 45.46  & 37.85 & 20.23 & 33.85 & \textbf{18.09} \\[4pt]

  Vicuna-13B  & 54.68  & 47.33  & 70.00 & 51.73 & 52.04 & 41.42 & 45.40 & \textbf{26.27}\\[4pt]
  
 WizardLM-13B & 48.60  & 34.75  & 56.38 & 45.86 & 57.56 & 22.42 & 29.07 & \textbf{15.87}  \\[4pt]

\revise{InternLM-7B}  & 45.72  &  37.23 & 65.12 &  41.49 & 42.33 & 25.23 & 29.38 & \textbf{10.45} \\[4pt]

InternLM-20B  & 59.14  &  51.05 & 68.75 &  53.47 & 62.35 & 34.99 & 47.06 & \textbf{29.52} \\[4pt]

Falcon-7B &  31.66 &  70.86 &  \phantom{0}3.77 & 10.52 & 14.85 & \phantom{0}7.69 & 14.38 & \phantom{0}\textbf{2.49} \\[4pt]
   
MPT-7B   &  35.60 &  \phantom{0}0.82 & 75.35 & 34.72 & \phantom{0}2.03 & \phantom{0}2.44 & 21.62 & \phantom{0}\textbf{3.52}\\[4pt]

\revise{GPT-3.5-turbo}  & 64.81 &  65.84  &  67.77  & 73.81  &  56.55 & 58.21 & 63.99 & \textbf{40.39} \\
\bottomrule
\end{tabular}
}
\end{center}
\label{table:LLMs-pos}
\end{table*}

\subsection{Symbol-Content Spurious Correlation}
In this subsection, we investigate the potential existence of shortcut correlations between option symbols and content. We ask whether the permutation required to fool the model’s prediction is independent of the chosen symbols or not. Here we consider two sets of symbols, Capital letters vs. Lowercase letters, and Capital letters vs. Roman Numerals.
For each set of symbols, we calculated the correlations across all permutations of test set predictions. Specifically, we examined the similarity in response patterns to permutations between different symbol sets, with a high correlation indicating similar responses to permutations across two sets, and a low correlation indicating the opposite.

\begin{figure}[h]
    \centering
    \includegraphics[width=\linewidth]{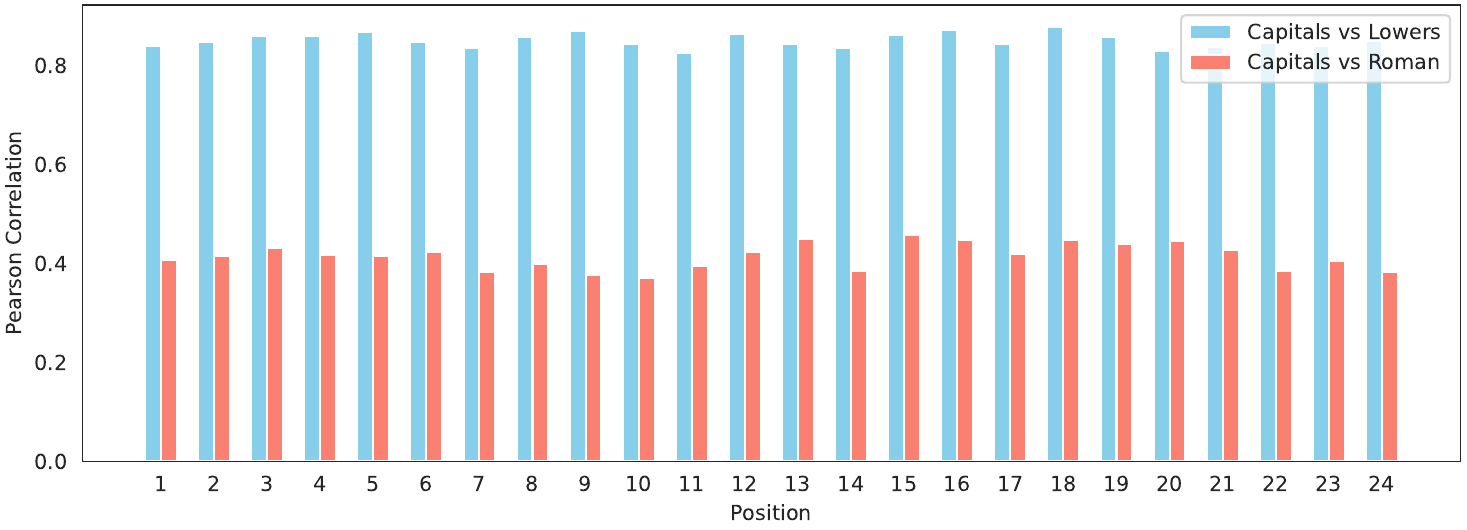}
    \caption{The correlation analysis of Llama2-13B model's predictions across different pairs of options symbols of each permutation reveals a notable finding: the low correlation score between permutation predictions when using capital letters and Roman numerals suggests that the model may have learned shortcuts or spurious correlations linking option symbols with answer content.}
    \label{fig:correlation}
\end{figure}

Our findings reveal a notably low correlation between the sets of capital letters (A/B/C/D) and Roman numerals (I/II/III/IV), in stark contrast to the correlation observed between capital letters (A/B/C/D) and lowercase letters (a/b/c/d). This discrepancy suggests that, while the baseline and permuted accuracies remain largely consistent across different symbol sets, their responses to permutations diverge significantly. Such behavior implies that the model might be exploiting symbol-answer shortcuts~\citep{geirhos2020shortcut,du2023shortcut} and spurious correlations~\citep{sagawa2020investigation} inadvertently learned during training, indicating another potential underlying cause of our observed vulnerability. Figure~\ref{fig:correlation} illustrates these dynamics, showcasing the Llama2-13B model's predictions and correlation across various symbol sets. 

In summary: not only are the models not invariant to permutation, but the specific way in which they are not invariant is dependent on the choice of symbols.

\begin{table}[h]
\begin{center}
    
\caption{Comparisons of Pearson correlation scores of different symbol sets on ARC-Challenge dataset averaged over different permutations.}
\resizebox{\linewidth}{!}{
\begin{tabular}{lccc}
\toprule
Symbol Set & Correlation & Original Acc. & Permuted Acc. \\
\midrule
Capital vs. Lowercase & 0.76 &  55.06 vs. 54.87  &  23.73 vs. 21.68  \\[4pt]
Capital vs. Roman & 0.36 &  55.06 vs. 52.49 &  23.73 vs. 19.33  \\
\bottomrule
\end{tabular}
}
\end{center}
\label{tab:correlation}
\end{table}

\section{Exploring Mitigation Strategies} %
The previous analysis of adversarial permutation vulnerability should be concerning to stakeholders interested in trustworthy and reliable AI, and suggests a new focus for researchers in developing models with improved intrinsic permutation robustness. Nevertheless, one might ask whether any post-hoc engineering fixes or fine-tuning could alleviate this issue in practice for existing models. We explore this question in this section.

\subsection{Post-hoc Mitigation Strategies}
For post-hoc strategies, we consider three strategies that have previously proven effective in improving model performance, namely, majority voting \citep{wang2023selfconsistency}, contextual calibration~\citep{zhao2021calibrate} and confidence-based voting, and ask whether they can alleviate adversarial permutation vulnerability.

\textbf{Setup}\quad 
Majority voting \citep{wang2023selfconsistency} has been shown highly successful in self-ensembling over stochastic predictions. In our context, we apply it by obtaining the predictions for all possible permutations and then selecting the most frequent prediction. If most permutations lead to a correct prediction and there are only one or two pathological permutations that lead to an incorrect prediction, then majority voting should provide complete robustness to adversarial permutation. 
Contextual calibration \citep{zhao2021calibrate} is designed to mitigate the prior bias introduced from the in-context examples by estimating the model’s bias toward each answer with a ``content-free'' query and fitting a calibration parameter. Here we consider the input question and options as the language prior bias. We first feed the model with content-free options (e.g. ``N/A'') as the content-free input, and then calibrate the real prediction based on the calibration parameters calculated from the content-free input. Additionally, we also apply confidence voting by taking the output that has maximum confidence among all permutations as the final prediction (M-confidence).

\textbf{Results}\quad 
From the results in Table~\ref{table:maj-cbu} for LLMs, we can see that neither defense proved effective at restoring the original performance levels. \revise{The majority voting and M-confidence certainly ameliorated the permutation attack as expected, but still fell short of the baseline accuracy with only very few models gaining improvement.} This is despite their being highly impractical defenses due to imposing a $k!$-fold increase in inference cost. We also additionally analyze the pattern of majority vote in Appendix Table~\ref{tab:maj_50percent} and~\ref{tab:maj_25percent}.
Contextual calibration, on the other hand, completely failed to make a meaningful impact on mitigating the adversarial attack. This re-confirms that the position bias is not the primary reason for models' permutation vulnerability. 

We additionally examine whether Pride~\cite{zheng2023sensitiveMCQ} can alleviate this issue, which introduces an inference-time debiasing method to mitigate token selection bias. Here, we debiased the prediction distribution of all possible permutations using their method and report original option order accuracy and permuted accuracy on the MMLU dataset in Table~\ref{tab:pride}. Although PriDe can indeed improve accuracy for the original order, it cannot effectively restore the performance with the permutation attack. It is understandable why PriDE almost completely fails: It learns a recalibration matrix that adjusts the relative probabilities of options (A, B, C, D). But this recalibration matrix is not input dependent, making it in the end a sophisticated amelioration against fixed positional/option bias. However, as we show here, the different permutations can lead to radically different preferences over options. In this case, no single fixed recalibration matrix can possibly alleviate this attack.

\begin{table*}[htb!]
\caption{Impact of majority vote, contextual calibration (C-Calibration), and maximum confidence (M-Confidence) defenses against the permutation attack on the MMLU dataset. Contextual calibration fails completely. Majority vote and M-Confidence ameliorate the attack, but do not completely restore performance. Red shading indicates below-chance results.}
\begin{center}
\resizebox{0.9\linewidth}{!}{%
\begin{tabular}{lllllll}
\toprule
{\bf Method} & {\bf Original} & {\bf Permutation Attack} & {\bf Majority Vote} & {\bf C-Calibration}  & {\bf \revise{M-Confidence}} \\
\midrule

Llama2-7B & 40.91 & \cellcolor{red!10}\phantom{0}6.17 (\textcolor{red}{34.74 $\downarrow$}) & 33.64 (\textcolor{red}{7.27 $\downarrow$}) & \cellcolor{red!10}\phantom{0}5.24 (\textcolor{red}{35.67 $\downarrow$}) & 22.62 \cellcolor{red!10}(\textcolor{red}{18.29 $\downarrow$}) \\[4pt]

Llama2-13B & 52.22 & \cellcolor{red!10}18.33 (\textcolor{red}{33.89 $\downarrow$}) & 48.53 (\textcolor{red}{3.69 $\downarrow$}) & \cellcolor{red!10}20.02 (\textcolor{red}{32.20 $\downarrow$}) & 50.83 \phantom{0}(\textcolor{red}{1.39 $\downarrow$})\\[4pt]

\revise{Llama2-70B} & 64.68 & 33.16 (\textcolor{red}{31.52 $\downarrow$}) & 65.37 (\textcolor{blue}{0.69 $\uparrow$}) & \cellcolor{red!10}35.77 (\textcolor{red}{28.91 $\downarrow$}) & 64.20 \phantom{0}(\textcolor{red}{0.48 $\downarrow$})\\[4pt]

Vicuna-v1.5-7B & 48.57 & \cellcolor{red!10}18.09 (\textcolor{red}{30.48 $\downarrow$}) & 44.10 (\textcolor{red}{4.47 $\downarrow$}) & \cellcolor{red!10}11.33 (\textcolor{red}{37.24 $\downarrow$}) & 38.29 (\textcolor{red}{10.28 $\downarrow$})\\[4pt]

Vicuna-v1.5-13B & 54.68 & 26.27 (\textcolor{red}{28.41 $\downarrow$}) & 52.03 (\textcolor{red}{2.65 $\downarrow$}) & \cellcolor{red!10}18.10 (\textcolor{red}{36.58 $\downarrow$}) & 55.58 \phantom{0}(\textcolor{blue}{0.90 $\uparrow$})\\[4pt]

WizardLM-13B & 48.60 & \cellcolor{red!10}15.87 (\textcolor{red}{32.73 $\downarrow$}) & 30.17 (\textcolor{red}{18.43 $\downarrow$}) & \cellcolor{red!10}\phantom{0}8.23 (\textcolor{red}{40.37 $\downarrow$}) & 37.81 (\textcolor{red}{11.21 $\downarrow$}) \\[4pt]

InternLM-20B & 59.14 & 29.52 (\textcolor{red}{29.62 $\downarrow$}) & 60.33 (\textcolor{blue}{1.19 $\uparrow$}) & 28.94 (\textcolor{red}{30.20 $\downarrow$}) & 64.80 \phantom{0}(\textcolor{blue}{5.66 $\uparrow$})\\[4pt]

Falcon-7b & 31.66 & \cellcolor{red!10}\phantom{0}2.49 (\textcolor{red}{29.17 $\downarrow$}) & \cellcolor{red!10}\phantom{0}4.38 (\textcolor{red}{27.28 $\downarrow$}) & \cellcolor{red!10}\phantom{0}3.59 (\textcolor{red}{28.07 $\downarrow$}) & \cellcolor{red!10}21.10 (\textcolor{red}{10.56 $\downarrow$}) \\[4pt]

MPT-7B & 35.60 & \cellcolor{red!10}\phantom{0}3.52 (\textcolor{red}{32.08 $\downarrow$})  & \cellcolor{red!10}13.80 (\textcolor{red}{21.80 $\downarrow$}) & \cellcolor{red!10}\phantom{0}6.24 (\textcolor{red}{29.36 $\downarrow$}) & \cellcolor{red!10}21.42 (\textcolor{red}{14.18 $\downarrow$})\\
\bottomrule
\end{tabular}
}
\end{center}
\label{table:maj-cbu}
\end{table*}

\begin{table}[ht]
\caption{Baseline and permuted accuracy with PriDe on MMLU. Pride cannot alleviate the vulnerability caused by full permutation.}
\begin{center}
\resizebox{\linewidth}{!}{
\begin{tabular}{lc}
\toprule
\textbf{Models} &  \textbf{PriDe (Baseline/Permuted Accuracy)} \\ 
\midrule
Llama2-7B &  42.60/\phantom{0}6.43 (\textcolor{red}{36.17$\downarrow$}) \\  
Llama2-13B &  52.41/19.26 (\textcolor{red}{33.15$\downarrow$}) \\  
Vicuna-7B &   49.37/18.42 (\textcolor{red}{30.95$\downarrow$}) \\  
Vicuna-13B &  55.27/27.68 (\textcolor{red}{27.59$\downarrow$}) \\ 
\bottomrule
\end{tabular}
}
\end{center}
\label{tab:pride}
\end{table}

\subsection{Fine-tuning on Training Set}
While (V)LLMs are typically evaluated in zero-shot on the MCQ benchmarks without within-dataset fine-tuning, we study whether fine-tuning on the training set can enhance the robustness to the permutation.

\textbf{Setup}\quad As many of the benchmarks do not provide a training set, we conduct two fine-tuning experiments using Llama2-7B on two datasets that do provide training sets: ARC-Challenge~\citep{clark2018arc} and MedMCQA~\cite{pal2022medmcqa}. We also consider two fine-tuning variations: using the original training data and augmenting the training data with option permutations ($n!$) during fine-tuning. The latter is inspired by adversarial training~\cite{madry2018towards}, to ensure that the model learns robustness to permutations during training. We fine-tune with LoRA~\cite{hu2021lora} for 1 epoch. 

\textbf{Results}\quad We present the results before and after fine-tuning in Table~\ref{tab:finetune}. We observe that fine-tuning indeed improves the baseline accuracy and the permuted accuracy. However, regular fine-tuning is not a very effective solution because the fine-tuned models still suffer substantially from permutation vulnerability, and not all datasets have training set for fine-tuning. On the other hand, fine-tuning with permutation, while does not reliably defeat the attack, substantially alleviates it compared to the zero-shot and regular fine-tuning baseline. Although this is not a universal solution as we expect the model to generalize in a zero-shot manner, this shows the potential of mitigating the sensitivity during the pre-training or supervised fine-tuning stage with more robust training strategies.

\begin{table}[hbt!]
\caption{Comparison of baseline and permuted accuracy for different fine-tuning strategies across two benchmarks with Llama2-7B. Fine-tuning with permutation can enhance the robustness to the permutation attacks compared to the zero-shot and regular fine-tuning baseline.}
\begin{center}
\resizebox{\linewidth}{!}{
\begin{tabular}{lcc}
\toprule
\textbf{Fine-tuning Strategy} & \textbf{ARC-Challenge} & \textbf{MedMCQA} \\ 
\midrule
\textbf{Zero-shot} & \cellcolor{red!10} 47.04/ 7.98 (\textcolor{red}{39.06$\downarrow$}) & \cellcolor{red!10} 37.96/ 1.60 (\textcolor{red}{36.36$\downarrow$}) \\  
\textbf{Regular Fine-tuning} & \cellcolor{red!10} 51.42/15.02 (\textcolor{red}{36.40$\downarrow$}) & \cellcolor{red!10} 45.03/14.38 (\textcolor{red}{30.65$\downarrow$}) \\
\textbf{Fine-tuning with Permutation} & 67.64/47.73 (\textcolor{red}{19.91$\downarrow$}) & 46.78/26.07 (\textcolor{red}{20.71$\downarrow$}) \\ 
\bottomrule
\end{tabular}
}
\end{center}
\label{tab:finetune}
\end{table}

\section{Related Work}
\keypoint{Large Language Models and Vision-Language Models.}
In recent years, the natural language processing community has seen astonishing progress in large language models (LLMs) with billions of trained parameters, such as GPT-3~\citep{brown2020language} and Llama~\citep{touvron2023llama, touvron2023llama2}, and become more intelligent after instruction-following fine-tuning~\citep{ouyang2022training, zheng2023vicuna}. With the strong capabilities of LLMs, there is a growing interest in grounding vision with LLMs to enable the models to perceive multimodal information~\citep{yin2023survey, zong2023SSML, li2023blip}, usually by utilizing pretrained language and vision encoders with trainable alignment modules to connect them. Such models have shown strong capabilities across a diverse range of tasks including multimodal generation, question-answering, dialogue, and more.

\keypoint{Multiple-Choice Question Answering (MCQA).}
Multiple-Choice Question Answering (MCQA) requires selecting the correct option from a set of choices and is prevalent in numerous real-world applications, making it a key performance metric for both LLMs and VLLMs. Various benchmarks such as MMLU~\citep{hendrycks2020measuring}, AGI-Eval~\citep{zhong2023agieval}, MedMCQA~\citep{pal2022medmcqa}, and SocialIQA~\citep{sapetal2019social} have been designed to assess MCQA proficiency across different domains. Different prompting approaches approaches have been considered for MCQA with multiple-choice prompting being the currently recommended state of the art \citep{robinson2023leveraging}. 
On these benchmarks, LLMs and VLLMs frequently achieve, or even surpass, human-level accuracy~\citep{anil2023palm, openai2023gpt4}, suggesting a high degree of reliability and robustness. However, we cast doubt on this presumed robustness, exposing the underlying fragility of these models in MCQA scenarios.

\keypoint{Robustness of LLMs and VLLMs.}
Despite their impressive capabilities, concerns remain about the robustness and reliability of LLMs and VLLMs~\citep{liu2023trustworthy}. Previous studies have revealed the sensitivity of LLMs to various factors including prompt~\citep{zhu2023promptbench}, in-context examples~\citep{liu2021makes, zhao2021calibrate}, irrelevant context~\citep{shi2023large}, etc. Despite its significance, the robustness of MCQA has been relatively unexamined, particularly for VLLMs. Our research addresses this gap by scrutinizing a specific, yet pervasive, vulnerability to answer choice permutations in MCQA across both model types. Concurrent work \citep{zheng2023sensitiveMCQ}  discusses position-bias in MCQA \revise{and \cite{liu2023mmbench} proposes circular evaluation}. Our results show that adversarial permutation vulnerability is a much deeper problem than position bias, and position calibration strategy (e.g.,~\cite{zheng2023sensitiveMCQ}) cannot solve this issue. %

\section{Discussion}
In this paper, we present a comprehensive empirical analysis that unveils a critical but often overlooked vulnerability in both large language models (LLMs) and large vision-language models (VLLMs) in the context of multiple-choice question answering (MCQA). Despite their seemingly robust performance on established MCQA benchmarks, these models are highly susceptible to simple manipulations like option permutations. Our findings raise concerns about the widespread practice of evaluating and deploying these models based on MCQA tasks, urging caution in interpreting high benchmark scores as evidence of robust capabilities.
We highlight the need for future work to develop training strategies and/or architectures that lead to intrinsic robustness to such adversarial attacks and develop parameter-efficient tuning approaches that can fine-tune or align existing pretrained LLMs and VLLMs to be invariant to permutations. Notably, although we do not offer a solution, we believe that our findings will be of interest to many researchers in the field and spur future works. Furthermore, these insights are crucial for industry professionals, particularly when deploying Large Language Models (LLMs) and Vision-Language Models (VLLMs) in real-world scenarios.

\section*{Impact Statement}
As a large number of LLM and VLLM benchmarks are based on multiple-choice question answering, the finding in our paper, that the MCQ evaluation is highly unstable and can be easily brought to below chance level through very simple permutations, is like to have a high impact on encouraging researchers to rethink the evaluation protocols and re-assess the capabilities of the developed LLMs and VLLMs. 

\section*{Acknowledgement}
Yongshuo Zong is supported by the United Kingdom Research and Innovation (grant EP/S02431X/1), UKRI Centre for Doctoral Training in Biomedical AI at the University of Edinburgh, School of Informatics. For the purpose of open access, the author has applied a creative commons attribution (CC BY) licence to any author accepted manuscript version arising.

\bibliography{iclr_ref}
\bibliographystyle{icml2024}

\newpage

\appendix
\onecolumn
\section{Appendix}

\subsection{Additional Results on Answer Set Pruning}
We present the additional results of the answer set pruning of all language and vision-language datasets in Table~\ref{tab:prune-arcc} to~\ref{tab:prune-mmbench}.

\subsection{Additional Results on Different Prompting and Attack Strategies}
\label{app:prompting}
In this subsection, we investigate the effect of different prompting techniques on the models' vulnerability to MCQA. The findings are summarized below.
\begin{itemize}
    \item  Table~\ref{tab:avg_permutation} presents the average number of permutations needed to break the predictions of ARC-Challenge and MedMCQA datasets. Observations are two fold: (1) stronger LLMs require a larger number of permutations to break, and (2) LLMs are easier to break on more difficult datasets (ARC-Challenge is relatively easier than MedMCQA for LLMs).
    \item Table~\ref{tab:maj_50percent} and~\ref{tab:maj_25percent} analyze the reason for the seemingly counterintuitive ineffectiveness of the majority vote with Llama2-7B (decrease with majority vote) and InternLM-20B (increase with majority vote). The intuitive summary is as follows (consistent with~\cite{chen2024more}): For easy queries where an LLM call's output is correct more than 50\% of the time, the probability of a correct majority vote goes 1 with infinite LLM calls. Conversely, for hard queries (e.g., with less than a 50\% correctness rate), the majority vote's accuracy trends towards 0 as LLM calls increase. 
    \item \revise{Table~\ref{table:LLMs-ICL} compares the performance before and after adversarial attack with in-context learning. Although in-context learning can improve the original performance, the models still suffer substantially from the performance drop after adversarial permutations. }
    \item  \revise{Table~\ref{table:mmlu-ICL} to Table~\ref{table:medmcqa-ICL} presents different attack strategies with in-context learning, i.e. permutation of in-context examples and searching for worst-case in-context examples. While they can decrease the performance, our adversarial attack has the biggest impact on the final performance and causes the largest performance drop. }
    \item  \revise{Table~\ref{tab:temperatures} compares different sampling strategies and temperatures. The performance of the other decoding strategies is even worse before and after the permutations compared to the greedy decoding we adopted. Therefore we can ensure our experiments were conducted properly and the findings can generalize to other decoding strategies.}
    \item  \revise{Table~\ref{tab:icl-vicuna13b} to Table~\ref{tab:icl-internlm20b} presents the effect of in-context learning on the position bias.  Our findings indicate that while the original model exhibited a preference for option B, this preference persisted even after introducing in-context examples with answers set to positions A, B, C, and D. This suggests that while in-context examples can modify the distribution across various options, they do not entirely override the inherent position bias of the model.}

\end{itemize}

\subsection{Further Analysis on Vision-Language Dataset}
\revise{We present analysis on vision-language dataset A-OKVQA~\citep{schwenk2022okvqa} in Table~\ref{table:VLLMs-pos} and~\ref{table:maj-cbu-vllm} about position bias and different strategies for mitigation. The additional analysis further ensures that our findings on LLMs can also be generalized to VLLMs.}

\subsection{Qualitative Results}
\label{app:qualitative}
\begin{figure}[h]
    \centering
    \includegraphics[width=\linewidth]{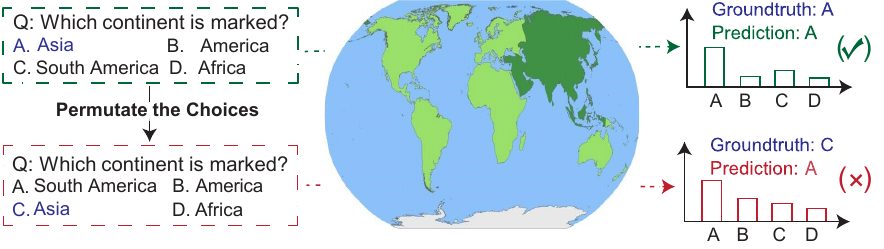}
    \caption{Qualitative results of permutations of answer options and the corresponding model (Otter-Llama) predictions. The example is selected from the ScienceQA dataset.}
    \label{fig:qua}
\end{figure}

To illustrate the permutation attack, we present qualitative results for LLMs in Table~\ref{table:lqa} and VLLMs in Fig.~\ref{fig:qua}.

\textbf{Language Models}\quad 
In Table~\ref{table:lqa}, we showcase an MCQA example from the ARC-challenge dataset~\citep{clark2018arc}, with the original answer order alongside two permutations. The ground-truth answer is underlined in each configuration. We use Llama-13B for this experiment. The model gives the correct prediction for the original option order. For permutation 1, if we only swap the position of option C and D, i.e., moving the ground-truth position to C, the model can still successfully give the prediction. However, for permutation 2, even if we do not move the ground-truth answer but only swap option A and B, the model incorrectly predicts A as the answer. This qualitative example underscores that the model's vulnerability extends beyond mere positional bias and even minor changes in option ordering can result in completely different predictions.

\textbf{Vision-Language Models}\quad 
In Appendix Figure~\ref{fig:qua}, we present a visual MCQA example from ScienceQA dataset using Otter-Llama model. In this example, we simply move the ground truth ``Asia'' from option A to option C. However, the model still predicts the answer to be A and shows strong confidence in terms of the token probabilities (right part of the figure). This might show the model's preference for the first option as a recency bias. 

\begin{figure*}
    \centering
    \includegraphics[width=0.9\textwidth]{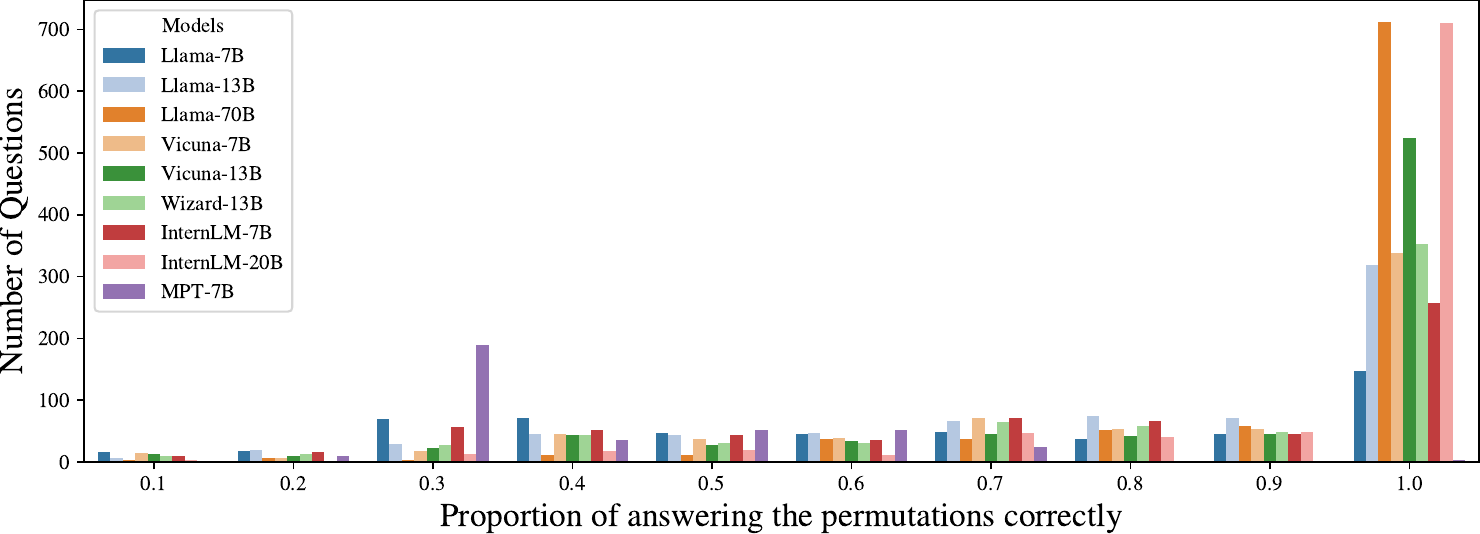}
    \vspace{-8pt}
    \caption{Analysis on permutation distribution. The histogram shows the number of questions for which the corresponding proportion of permutations leads to the correct answer (ideal is a full bar at the 100\% bin, indicating that all permutations are correctly answered for all questions). The distribution of bins suggests that many questions have multiple adversarial permutations.}
    \label{fig:histogram}
\end{figure*}

\textbf{Analysis on Permutation Distribution}\quad
While our main focus has been on the permutation-robustness of LLMs and VLMs, we can also ask about the distribution of responses as a function of permutation. For example, is there only one specific pathological permutation among all $k!$ options, or are there many mistake-inducing permutations? To analyse this we report in Figure~\ref{fig:histogram}, a histogram over the questions in ARC-challenge where each bin represents the number of questions where the specified proportion of permutations led to the correct answer that are originally correctly answered. For example, we see that Llama2-70B has a large number of questions that succeed for {almost} all permutations, while several models have a substantial batch of questions that are only correctly answered for around 30\% of the potential permutations. Interestingly, most models have a substantial minority of questions that are only correctly answered for a small fraction of the permutations (leftmost bin).%

\newpage

\begin{table*}[h]
    \caption{Results of answer Set Pruning on ARC (challenge) Dataset.}
    \begin{center}
        \begin{tabular}{lllllll}
            \toprule
    {\bf Method} & {\bf Original 4 Choices} & {\bf 3 Choices} & {\bf 2 Choices} \\
    \midrule
    Llama2-7B & 47.04 & \cellcolor{red!10}52.13/25.67 (\textcolor{red}{{26.46$\downarrow$}}) & \cellcolor{red!10}69.44/27.04 (\textcolor{red}{{42.40$\downarrow$}}) \\[4pt]
    Llama2-13B & 61.80 & \cellcolor{red!10}68.07/27.55 (\textcolor{red}{{40.52$\downarrow$}}) & \cellcolor{red!10}77.08/39.57 (\textcolor{red}{{37.51$\downarrow$}}) \\[4pt]
    \revise{Llama2-70B} & 80.00 & 83.09/45.88 (\textcolor{red}{{37.21$\downarrow$}}) &
    84.21/66.01 (\textcolor{red}{{18.20$\downarrow$}})\\[4pt]
    Vicuna-v1.5-7B & 58.37 & \cellcolor{red!10}68.07/29.70 (\textcolor{red}{{38.37$\downarrow$}}) & \cellcolor{red!10}78.11/42.66 (\textcolor{red}{{35.45$\downarrow$}}) \\[4pt]
    Vicuna-v1.5-13B & 69.27 & 74.85/42.32 (\textcolor{red}{{32.53$\downarrow$}}) & 83.18/56.14 (\textcolor{red}{{27.04$\downarrow$}}) \\[4pt]
    WizardLM-13B & 58.20 & \cellcolor{red!10}67.38/28.07 (\textcolor{red}{{39.31$\downarrow$}}) & \cellcolor{red!10}76.05/4.64 (\textcolor{red}{{71.41$\downarrow$}}) \\[4pt]
    \revise{InternLM-7B} & 56.14 & \cellcolor{red!10}61.37/17.42 (\textcolor{red}{43.95$\downarrow$}) & \cellcolor{red!10}71.93/29.44 (\textcolor{red}{42.49$\downarrow$}) \\[4pt]
    InternLM-20B & 78.28 & 82.06/48.58 (\textcolor{red}{{33.48$\downarrow$}}) & 84.81/56.03 (\textcolor{red}{{28.78$\downarrow$}}) \\[4pt]
    Falcon-7B & 34.74 & \cellcolor{red!10}31.76/0.00 (\textcolor{red}{{31.76$\downarrow$}}) & \cellcolor{red!10}48.58/0.43 (\textcolor{red}{{48.15$\downarrow$}}) \\[4pt]
    MPT-7B & 37.76 & \cellcolor{red!10}40.43/12.15 (\textcolor{red}{{28.28$\downarrow$}}) & \cellcolor{red!10}50.47/0.09 (\textcolor{red}{{50.38$\downarrow$}}) \\[4pt]
    Random Chance & 25.0 & 33.33 & 50.0 \\ 
\bottomrule
        \end{tabular}
    \end{center}
    \label{tab:prune-arcc}
\end{table*}

\begin{table*}
    \caption{Results of answer set pruning on SocialiQA Dataset.}
    \begin{center}
        \begin{tabular}{lllllll}
        \toprule
            {\bf Method} & {\bf Original 4 Choices} & {\bf 2 Choices} \\
\midrule
Llama2-7B & 52.00 & \cellcolor{red!10}68.42/29.58 (\textcolor{red}{{38.84$\downarrow$}}) \\[4pt]
Llama2-13B & 61.21 & \cellcolor{red!10}73.64/45.91 (\textcolor{red}{{27.73$\downarrow$}}) \\[4pt]
\revise{Llama2-70B} & 71.60 & 81.88/53.99 (\textcolor{red}{{27.89$\downarrow$}}) \\[4pt]
Vicuna-v1.5-7B & 64.99 & \cellcolor{red!10}73.29/41.56 (\textcolor{red}{{31.73$\downarrow$}}) \\[4pt]
Vicuna-v1.5-13B & 66.07 & 78.25/53.48 (\textcolor{red}{{24.77$\downarrow$}}) \\[4pt]
WizardLM-13B & 79.48 & \cellcolor{red!10}69.75/30.91 (\textcolor{red}{{38.84$\downarrow$}}) \\[4pt]
\revise{InternLM-7B} & 59.47 &  76.41/53.02 (\textcolor{red}{23.39$\downarrow$})\\[4pt]
InternLM-20B & 36.29 & 86.8/72.82 (\textcolor{red}{{13.98$\downarrow$}}) \\[4pt]
Falcon-7B & 41.61 & \cellcolor{red!10}55.83/0.85 (\textcolor{red}{{54.98$\downarrow$}}) \\[4pt]
MPT-7B & 70.62 & \cellcolor{red!10}60.08/4.52 (\textcolor{red}{{55.56$\downarrow$}}) \\[4pt]
Random Chance & 25.0 & 50.0 \\ 
\bottomrule
        \end{tabular}
    \end{center}
    \label{tab:prune-socialiqa}
\end{table*}

\begin{table*}
    \caption{Results of answer set Pruning on MedMCQ Dataset.}
    \begin{center}
        \begin{tabular}{lclllll}
            \toprule
    {\bf Method} & {\bf Original 4 Choices} & {\bf 3 Choices} & {\bf 2 Choices} \\
    \midrule
    Llama2-7B & 37.96 & \cellcolor{red!10}47.94/2.63 (\textcolor{red}{{45.31$\downarrow$}}) & \cellcolor{red!10}62.29/5.68 (\textcolor{red}{{56.61$\downarrow$}}) \\[4pt]
    Llama2-13B & 39.78 & \cellcolor{red!10}50.02/19.35 (\textcolor{red}{{30.67$\downarrow$}}) & \cellcolor{red!10}10.37/37.45 (\textcolor{red}{{-27.08$\downarrow$}}) \\[4pt]
    \revise{Llama2-70B} & 49.61 & \cellcolor{red!10}55.33/17.86 (\textcolor{red}{{37.47$\downarrow$}}) & \cellcolor{red!10}65.66/28.76 (\textcolor{red}{{36.90$\downarrow$}})  \\[4pt]
    
    Vicuna-v1.5-7B & 39.28 & \cellcolor{red!10}46.27/10.55 (\textcolor{red}{{35.72$\downarrow$}}) & \cellcolor{red!10}59.62/22.13 (\textcolor{red}{{37.49$\downarrow$}}) \\[4pt]
    Vicuna-v1.5-13B & 41.80 & \cellcolor{red!10}50.33/24.72 (\textcolor{red}{{25.61$\downarrow$}}) & \cellcolor{red!10}61.23/28.70 (\textcolor{red}{{32.53$\downarrow$}}) \\[4pt]
    WizardLM-13B & 34.87 & \cellcolor{red!10}40.39/8.37 (\textcolor{red}{{32.02$\downarrow$}}) & \cellcolor{red!10}52.46/10.79 (\textcolor{red}{{41.67$\downarrow$}}) \\[4pt]
    \revise{InternLM-7B} & 56.14 & \cellcolor{red!10}38.49/2.66 (\textcolor{red}{35.83$\downarrow$}) & \cellcolor{red!10}52.02/8.42 (\textcolor{red}{43.60$\downarrow$}) \\[4pt]
    InternLM-20B & 43.61 & \cellcolor{red!10}60.05/17.59 (\textcolor{red}{{42.46$\downarrow$}}) & \cellcolor{red!10}65.08/30.61 (\textcolor{red}{{34.47$\downarrow$}}) \\[4pt]
    Falcon-7B & 28.12 & \cellcolor{red!10}36.18/1.05 (\textcolor{red}{{35.13$\downarrow$}}) & \cellcolor{red!10}51.30/5.23 (\textcolor{red}{{46.07$\downarrow$}}) \\[4pt]
    MPT-7B & 26.31 & \cellcolor{red!10}38.33/3.07 (\textcolor{red}{{35.26$\downarrow$}}) & \cellcolor{red!10}53.50/9.71 (\textcolor{red}{{43.79$\downarrow$}}) \\[4pt]
    Random Chance & 25.0 & 33.33 & 50.0 \\ 
    \bottomrule
        \end{tabular}
    \end{center}
    \label{tab:prune-medmcq}
\end{table*}

\begin{table*}
    \caption{Results of answer set pruning on A-OKVQA dataset. Numbers in each cell represent original accuracy, accuracy after adversarial permutation attack, and relative performance drop.}
    \begin{center}
        \begin{tabular}{lcll}
\toprule
{\bf Method} & {\bf 4 Choices} & {\bf 3 Choices} & {\bf 2 Choices} \\
\midrule
InstructBLIP7B & 74.06 & 79.21/45.92 (\textcolor{red}{{33.29$\downarrow$}}) & 85.07/54.85 (\textcolor{red}{{30.22$\downarrow$}}) \\[4pt]
InstructBLIP13B & 77.90 & 81.66/48.33 (\textcolor{red}{{33.33$\downarrow$}}) & 88.56/56.52 (\textcolor{red}{{32.04$\downarrow$}}) \\[4pt]
OpenFlamingo & 46.90 & \cellcolor{red!10}54.18/4.88 (\textcolor{red}{{49.30$\downarrow$}}) & \cellcolor{red!10}66.90/5.09 (\textcolor{red}{{61.81$\downarrow$}}) \\[4pt]
Otter-Llama7B & 57.99 & \cellcolor{red!10}64.98/33.10 (\textcolor{red}{{31.88$\downarrow$}}) & 75.02/39.74 (\textcolor{red}{{35.28$\downarrow$}}) \\[4pt]
Otter-MPT7B & 68.21 & \cellcolor{red!10}76.16/46.11 (\textcolor{red}{{30.05$\downarrow$}}) &\cellcolor{red!10}81.48/51.44 (\textcolor{red}{{30.04$\downarrow$}}) \\[4pt]
Llava-7B & 52.91 & 42.86/9.44 (\textcolor{red}{{33.42$\downarrow$}}) & \cellcolor{red!10}63.55/12.90 (\textcolor{red}{{50.65$\downarrow$}}) \\[4pt]
Llava-13B & 63.14 & 71.09/33.37 (\textcolor{red}{{37.72$\downarrow$}}) & 76.24/41.22 (\textcolor{red}{{35.02$\downarrow$}}) \\[4pt]
Limber & 39.91 & \cellcolor{red!10}49.69/4.54 (\textcolor{red}{{45.15$\downarrow$}}) & \cellcolor{red!10}65.68/18.08 (\textcolor{red}{{47.60$\downarrow$}}) \\[4pt]
mPLUG-Owl-pt & 39.91 & \cellcolor{red!10}45.59/4.95 (\textcolor{red}{{40.64$\downarrow$}}) & \cellcolor{red!10}56.42/10.57 (\textcolor{red}{{45.85$\downarrow$}}) \\[4pt]
mPLUG-Owl-instr & 37.12 & \cellcolor{red!10}47.86/5.15 (\textcolor{red}{{42.71$\downarrow$}}) & \cellcolor{red!10}58.92/16.77 (\textcolor{red}{{42.15$\downarrow$}}) \\[4pt]
Random Chance & 25.0 & 33.33 & 50.0 \\ 
\bottomrule
        \end{tabular}
        \end{center}
    \label{tab:prune-aokvqa}
\end{table*}

\begin{table*}
    \caption{Results of answer set pruning on SEED-Bench dataset. Numbers in each cell represent original accuracy, accuracy after adversarial permutation attack, and relative performance drop.}
    \begin{center}
        \begin{tabular}{lcll}
            \toprule
            {\bf Model} & {\bf Original 4 Choices} & {\bf 3 Choices} & {\bf 2 Choices} \\
            \midrule
            InstructBLIP7B & 51.61 & 59.83/38.12(\textcolor{red}{{$21.71\downarrow$}}) & 70.69/55.62(\textcolor{red}{{$15.07\downarrow$}}) \\[4pt]
            InstructBLIP13B & 53.65 & 61.22/42.97(\textcolor{red}{{$18.25\downarrow$}}) & 72.79/57.18(\textcolor{red}{{$15.61\downarrow$}}) \\[4pt]
            OpenFlamingo & 37.99 & 39.64/\cellcolor{red!10}10.25(\textcolor{red}{{$29.39\downarrow$}}) & \cellcolor{red!10}55.31/28.35(\textcolor{red}{{$26.96\downarrow$}}) \\[4pt]
            Otter-Llama7B & 50.77 & 49.46/\cellcolor{red!10}13.48(\textcolor{red}{{35.98$\downarrow$}}) & 63.57/\cellcolor{red!10}18.86(\textcolor{red}{{$44.71\downarrow$}}) \\[4pt]
            Otter-MPT7B & 46.76 & 54.18/\cellcolor{red!10}17.42(\textcolor{red}{{$36.76\downarrow$}}) & \cellcolor{red!10}66.43/28.35(\textcolor{red}{{$38.08\downarrow$}}) \\[4pt]
            Llava-7B & 38.36 & 43.30/\cellcolor{red!10}6.11(\textcolor{red}{{$37.19\downarrow$}}) & 57.50/\cellcolor{red!10}7.48(\textcolor{red}{{$50.02\downarrow$}}) \\[4pt]
            Llava-13B & 44.00 & 52.51/\cellcolor{red!10}17.25(\textcolor{red}{{$35.26\downarrow$}}) & \cellcolor{red!10}63.37/25.24(\textcolor{red}{{$38.13\downarrow$}}) \\[4pt]
            Limber & 31.50 & \cellcolor{red!10}38.76/1.13(\textcolor{red}{{$37.63\downarrow$}}) & \cellcolor{red!10}55.58/0.06(\textcolor{red}{{$55.52\downarrow$}}) \\[4pt]
            mPLUG-Owl-pt & 35.57 & 42.08/\cellcolor{red!10}1.79(\textcolor{red}{{$40.29\downarrow$}}) & 57.72/\cellcolor{red!10}3.88(\textcolor{red}{{$53.84\downarrow$}}) \\[4pt]
            mPLUG-Owl-instr & 36.74 & 44.35/\cellcolor{red!10}2.94(\textcolor{red}{{$41.41\downarrow$}}) & 56.61/\cellcolor{red!10}7.53(\textcolor{red}{{$49.08\downarrow$}}) \\[4pt]
            Random Chance & 25.0 & 33.33 & 50.0 \\
            \toprule
        \end{tabular}
        \end{center}
    \label{tab:prune-seedbench}
\end{table*}

\begin{table*}
    \caption{Results of answer set pruning on MMBench dataset. Numbers in each represent original accuracy, accuracy after adversarial attack, and relative performance drop.}
    \begin{center}
        \begin{tabular}{lcll}
            \toprule
            {\bf Method} & {\bf Original 4 Choices} & {\bf 3 Choices} & {\bf 2 Choices} \\[4pt]
            \midrule
            InstructBLIP7B & 64.91 & 72.61/45.28(\textcolor{red}{{$27.33\downarrow$}}) & 79.62/51.15(\textcolor{red}{{$28.47\downarrow$}}) \\[4pt]
            InstructBLIP13B & 67.12 & 72.79/50.42(\textcolor{red}{{$22.37\downarrow$}}) & 81.27/57.30(\textcolor{red}{{$23.97\downarrow$}}) \\[4pt]
            OpenFlamingo & 38.99 & 46.58/\cellcolor{red!10}3.81(\textcolor{red}{{$42.77\downarrow$}}) & 59.65/\cellcolor{red!10}7.42(\textcolor{red}{{$52.23\downarrow$}}) \\[4pt]
            Otter-Llama7B & 55.24 & 61.73/25.11(\textcolor{red}{{$36.62\downarrow$}}) & 73.02/32.99(\textcolor{red}{{$40.03\downarrow$}}) \\[4pt]
            Otter-MPT7B & 61.31 & 66.71/28.53(\textcolor{red}{{$38.18\downarrow$}}) & 75.28/46.06(\textcolor{red}{{$29.22\downarrow$}}) \\[4pt]
            Llava-7B & 46.03 & 45.37/\cellcolor{red!10}2.14(\textcolor{red}{{$43.23\downarrow$}}) & 59.42/\cellcolor{red!10}3.66(\textcolor{red}{{$55.76\downarrow$}}) \\[4pt]
            Llava-13B & 59.13 & 65.20/35.66(\textcolor{red}{{$29.54\downarrow$}}) & 73.13/42.49(\textcolor{red}{{$30.64\downarrow$}}) \\[4pt]
            Limber & 34.93 & \cellcolor{red!10}44.23/2.56(\textcolor{red}{{$41.67\downarrow$}}) & \cellcolor{red!10}61.00/12.75(\textcolor{red}{{$48.25\downarrow$}}) \\[4pt]
            mPLUG-Owl-pt & 42.57 & 53.17/\cellcolor{red!10}12.24(\textcolor{red}{{$40.93\downarrow$}}) & 56.42/\cellcolor{red!10}16.90(\textcolor{red}{{$39.52\downarrow$}}) \\[4pt]
            mPLUG-Owl-instr & 43.74 & 50.17/\cellcolor{red!10}11.15(\textcolor{red}{{$39.02\downarrow$}}) & 61.48/\cellcolor{red!10}18.97(\textcolor{red}{{$42.51\downarrow$}}) \\[4pt]
            Random Chance & 25.0 & 33.33 & 50.0 \\
            \toprule
        \end{tabular}
        \end{center}
    \label{tab:prune-mmbench}
    
\end{table*}

\begin{table*}[h]
\caption{Average number of permutations needed to break the predictions of ARC-Challenge and MedMCQA datasets. In each cell, the first number is the average number of all samples, and the second number is the average number of permutations needed when the question in the original order is answered correctly. }
\begin{center}
\resizebox{0.5\linewidth}{!}{%
\begin{tabular}{lll}
\toprule
\textbf{Models} & \textbf{ARC-C} & \textbf{MedMCQA} \\ 
\midrule
Llama2-7B & 3.73/7.96 & 2.01/5.26 \\  
Llama2-13B & 7.53/12.19 & 3.40/8.90 \\  
Llama2-70B & 13.80/17.16 & 5.46/11.10 \\  
Vicuna-v1.5 & 7.47/12.80 & 3.18/8.14 \\ 
Vicuna-v1.5-13B & 10.77/15.55 & 4.42/10.60 \\  
WizardLM-13B & 7.19/12.35 & 2.73/8.08 \\ 
InternLM-7B & 6.00/10.69 & 1.58/4.84 \\  
InternLM-20B & 14.19/18.13 & 4.72/10.82 \\  
Falcon-7b & 1.31/5.83 & 1.88/5.71 \\ 
MPT-7B & 1.59/2.87 & 0.58/2.18 \\ 
\bottomrule
\end{tabular}
}
\end{center}
\label{tab:avg_permutation}
\end{table*}

\begin{table*}[h]
\begin{center}
\caption{Majority vote performance when considering the easy sample with respect to a certain model to be the ones where 50\% of the permutations are correct. If we consider the easy sample with respect to a certain model to be the ones where 50\% of the permutations are correct, the majority vote accuracy of easy samples will be 100\%, and we calculate the accuracy of the majority of the difficult samples. For both Llama2-7B and InternLM-20B, the majority vote accuracies of difficult samples are very low. However, there are more easy samples wrt InternLM-20B compared to Llama2-7B, and therefore InternLM-20B overall improves with the majority vote while the accuracy of Llama2-7B decreases with the majority vote.}
\label{tab:maj_50percent}
\resizebox{\linewidth}{!}{%
\begin{tabular}{lccc}
\toprule
\textbf{Model} & \textbf{Performance (original/majority vote)} & \textbf{Easy sample Acc. (\# samples)} & \textbf{Difficult sample Acc. (\# samples)} \\
\midrule
Llama2-7B & 40.91/33.64 & 100 (4148) & 16.56 (9334) \\
InternLM-20B & 59.14/60.33 & 100 (9957) & 10.69 (3512) \\
\bottomrule
\end{tabular}
}
\end{center}
\end{table*}

\begin{table*}[h]
\begin{center}
\caption{Majority vote performance when considering the easy sample with respect to a certain model to be the ones where 25\% of the permutations are correct on the MMLU dataset. If we consider the easy sample with respect to a certain model to be the ones where 25\% of the permutations are correct, the majority vote accuracy of difficult samples will be 0\%, and we calculate the accuracy for majority vote on the easy samples. As shown in the table, both models have improved with majority vote on easy samples but the final performance may or may not be improved overall when considering the average accuracy across both easy and difficult samples.}
\label{tab:maj_25percent}
\resizebox{\linewidth}{!}{%
\begin{tabular}{lccc}
\toprule
\textbf{Model} & \textbf{Performance (original/majority vote)} & \textbf{Easy sample Acc. (\# samples)} & \textbf{Difficult sample Acc. (\# samples)} \\
\midrule
Llama2-7B & 40.91/33.64 & 46.08 (10252) & 0 (3790) \\
InternLM-20B & 59.14/60.33 & 75.28 (11530) & 0 (2512) \\
\bottomrule
\end{tabular}
}
\end{center}
\end{table*}

\begin{table*}[t]
\caption{Performance comparisons of LLMs before and after adversarial attack with in-context learning prompt. Numbers in each represent original accuracy, accuracy after adversarial attack, and relative performance drop. Red shading indicates experiments where the permutation attack reduced performance below chance level. All models suffer substantially with most experiments leading to below chance performance.}
\begin{center}
\resizebox{\linewidth}{!}{%
\begin{tabular}{llllllll}
\toprule
{\bf Method} & {\bf MMLU} & {\bf ARC-c} & {\bf BoolQ}  &  {\bf SocialiQA }& {\bf MedMCQA}\\
\midrule

Llama2-7B & \cellcolor{red!10}45.80/11.16 (\textcolor{red}{34.64 $\downarrow$}) & \cellcolor{red!10}47.04/\phantom{0}7.98 (\textcolor{red}{39.06 $\downarrow$}) & \cellcolor{red!10}61.79/\phantom{0}8.23 (\textcolor{red}{53.56 $\downarrow$}) &\cellcolor{red!10}52.00/15.71 (\textcolor{red}{36.29 $\downarrow$}) &\cellcolor{red!10}37.96/\phantom{0}1.60 (\textcolor{red}{36.36 $\downarrow$}) \\[4pt]

Llama2-13B & \cellcolor{red!10}55.37/21.69 (\textcolor{red}{33.68$\downarrow$}) & \cellcolor{red!10}61.80/21.63 (\textcolor{red}{40.17 $\downarrow$}) & \cellcolor{red!10}67.16/38.29 (\textcolor{red}{28.87 $\downarrow$}) & 61.21/34.14 (\textcolor{red}{27.07 $\downarrow$}) & \cellcolor{red!10}39.78/\phantom{0}7.35 (\textcolor{red}{32.43 $\downarrow$}) \\[4pt]

\revise{Llama2-70B} & 68.86/41.30 (\textcolor{red}{27.56$\downarrow$})  & 80.00/51.50 (\textcolor{red}{28.50 $\downarrow$}) & 76.39/56.21 (\textcolor{red}{20.18 $\downarrow$}) & 71.60/49.85 (\textcolor{red}{21.75 $\downarrow$}) & \cellcolor{red!10}49.61/\phantom{0}7.35 (\textcolor{red}{42.26 $\downarrow$}) \\[4pt]

Vicuna-v1.5 & \cellcolor{red!10}49.89/19.61 (\textcolor{red}{30.28$\downarrow$}) & \cellcolor{red!10}58.37/23.43 (\textcolor{red}{34.94 $\downarrow$}) & \cellcolor{red!10}64.04/29.60 (\textcolor{red}{34.44 $\downarrow$}) & 64.99/38.33 (\textcolor{red}{26.66 $\downarrow$}) & \cellcolor{red!10}39.28/\phantom{0}7.67 (\textcolor{red}{31.61 $\downarrow$}) \\[4pt]

Vicuna-v1.5-13B & 55.80/27.33 (\textcolor{red}{28.47$\downarrow$}) & 69.27/38.80 (\textcolor{red}{30.47 $\downarrow$}) &  \cellcolor{red!10}68.96/42.14 (\textcolor{red}{26.82 $\downarrow$}) & 66.07/44.42 (\textcolor{red}{21.65 $\downarrow$}) & \cellcolor{red!10}41.80/11.90 (\textcolor{red}{29.90 $\downarrow$}) \\[4pt]

WizardLM-13B & \cellcolor{red!10}48.93/14.40 (\textcolor{red}{34.53$\downarrow$})  & \cellcolor{red!10}58.20/21.12 (\textcolor{red}{37.08 $\downarrow$}) & \cellcolor{red!10}67.49/42.11 (\textcolor{red}{25.38 $\downarrow$}) & \cellcolor{red!10}63.46/31.78 (\textcolor{red}{31.68 $\downarrow$}) & \cellcolor{red!10}34.87/\phantom{1}6.32 (\textcolor{red}{28.55 $\downarrow$}) \\[4pt]

\revise{InternLM-7B} & \cellcolor{red!10}48.36/15.90 (\textcolor{red}{32.46$\downarrow$}) & \cellcolor{red!10}56.14/17.34 (\textcolor{red}{38.80
 $\downarrow$}) & \cellcolor{red!10}65.83/26.41 (\textcolor{red}{39.42 $\downarrow$}) & \cellcolor{red!10}59.47/30.30 (\textcolor{red}{29.17 $\downarrow$}) & \cellcolor{red!10}32.63/\phantom{1}2.56 (\textcolor{red}{30.07 $\downarrow$}) \\[4pt]

InternLM-20B & 60.50/32.14 (\textcolor{red}{28.36$\downarrow$}) & 78.28/54.42 (\textcolor{red}{23.86 $\downarrow$}) & 85.20/82.91 (\textcolor{red}{\phantom{0}2.29 $\downarrow$}) & 79.48/65.97 (\textcolor{red}{13.51 $\downarrow$}) & \cellcolor{red!10}43.61/13.92 (\textcolor{red}{29.69 $\downarrow$}) \\[4pt]

Falcon-7b & \cellcolor{red!10}26.95/\phantom{0}0.00 (\textcolor{red}{26.95$\downarrow$}) & \cellcolor{red!10}34.74/\phantom{0}0.09 (\textcolor{red}{34.65 $\downarrow$}) \cellcolor{red!10}&\cellcolor{red!10}55.35/\phantom{0}2.66 (\textcolor{red}{52.69 $\downarrow$}) & \cellcolor{red!10}36.29/\phantom{0}0.55 (\textcolor{red}{35.74 $\downarrow$}) & \cellcolor{red!10}28.12/\phantom{0}0.07 (\textcolor{red}{28.05 $\downarrow$}) \\[4pt]

MPT-7B & \cellcolor{red!10}38.73/\phantom{0}5.21 (\textcolor{red}{33.52 $\downarrow$}) & \cellcolor{red!10}37.76/\phantom{0}1.06 (\textcolor{red}{36.70 $\downarrow$}) &\cellcolor{red!10}58.46/\phantom{0}7.03 (\textcolor{red}{51.43 $\downarrow$}) & \cellcolor{red!10}41.61/\phantom{0}2.53 (\textcolor{red}{39.08 $\downarrow$}) & \cellcolor{red!10}26.31/\phantom{0}1.60 (\textcolor{red}{24.71 $\downarrow$}) \\[4pt]

Random Chance & 25.0  & 25.0  & 50.0 &  33.33 & 25.0 \\

\bottomrule
\end{tabular}
}
\end{center}
\label{table:LLMs-ICL}
\end{table*}

\begin{table*}[h]
\begin{center}
\caption{\revise{Comparisons of different attacks on the MMLU Dataset. In-context learning (ICL) improves the zero-shot performance, and attacks on in-context examples can decrease the performance. However, our adversarial attack has the biggest impact on the final performance (largest drop).}}

\resizebox{\linewidth}{!}{%
\begin{tabular}{lccccll}
\toprule
\textbf{Model} & \textbf{Original 0-shot} & \textbf{ICL} & \textbf{ICL Permutation} & \textbf{ICL Search} & \textbf{Permutation Attack} \\
\midrule
Llama2-7B & 40.91 & 45.80 & 35.09 & 34.46 & 6.17 \\[4pt]
Llama2-13B & 52.22 & 55.37 & 46.65 & 46.07 & 18.33 \\[4pt]
Llama2-70B & 64.68 & 68.86 & 59.82 & 59.68 & 33.16 \\[4pt]
Vicuna-v1.5 & 48.57 & 49.89 & 40.85 & 41.92 & 18.09 \\[4pt]
Vicuna-v1.5-13B & 54.68 & 55.80 & 54.65 & 49.17 & 26.27 \\[4pt]
WizardLM-13B & 48.60 & 48.93 & 39.98 & 48.27 & 15.87 \\[4pt]
InternLM-7B & 45.72 & 48.36 &  37.35 & 38.17 & 10.45 \\[4pt]
InternLM-20B & 59.14 & 60.50 & 54.94  & 54.45 & 29.52 \\[4pt]
Falcon-7b & 31.66 & 26.95 & 27.18 & 26.79 & 2.49 \\[4pt]
MPT-7B & 35.60 & 38.73 & 30.51 & 27.33 & 3.52 \\
\bottomrule
\end{tabular}
}
\label{table:mmlu-ICL}
\end{center}
\end{table*}

\begin{table*}[h]
\begin{center}
\caption{\revise{Comparisons of different attacks on the ARC-Challenge Dataset. In-context learning (ICL) improves the zero-shot performance, and the attack on in-context examples can decrease the performance. However, our adversarial attack has the biggest impact on the final performance (largest performance drop).}}
\resizebox{\linewidth}{!}{%
\begin{tabular}{lccccll}
\toprule
\textbf{Model} & \textbf{Original 0-shot} & \textbf{ICL} & \textbf{ICL Permutation} & \textbf{ICL Search} & \textbf{Permutation Attack} \\
\midrule
Llama2-7B & 47.04 & 54.16 & 45.75  & 38.20 & 7.98 \\[4pt]
Llama2-13B & 61.80 & 66.70 & 59.31 & 53.91 & 21.63 \\[4pt]
Llama2-70B & 80.00 & 84.55 & 80.00 & 79.23 & 51.5 \\[4pt]
Vicuna-v1.5 & 58.37 & 60.60 & 54.33  & 50.64 & 23.43 \\[4pt]
Vicuna-v1.5-13B & 69.27 &  72.02 &  66.44 & 60.52 & 38.8 \\[4pt]
WizardLM-13B & 58.20 & 59.74 & 49.01  & 43.26 & 21.12 \\[4pt]
InternLM-7B & 56.14 & 65.06 &  55.54 & 51.59 & 17.34 \\[4pt]
InternLM-20B & 78.28 & 80.52 &  76.74 & 74.33 & 54.42 \\[4pt]
Falcon-7b & 34.74 & 37.98 & 28.46 & 22.15 & 0.09 \\[4pt]
MPT-7B & 37.76 & 41.26 & 31.99 & 26.37 & 1.06 \\
\bottomrule
\end{tabular}
}
\label{table:arc-ICL}
\end{center}
\end{table*}

\begin{table}[h]
\begin{center}
\caption{\revise{Comparisons of different attacks on the BoolQ Dataset. In-context learning (ICL) improves the zero-shot performance, and the attack on in-context examples can decrease the performance. However, our adversarial attack has the biggest impact on the final performance (largest performance drop).}}
\resizebox{\linewidth}{!}{%
\begin{tabular}{lccccll}
\toprule
\textbf{Model} & \textbf{Original 0-shot} & \textbf{ICL} & \textbf{ICL Permutation} & \textbf{ICL Search} & \textbf{Permutation Attack} \\
\midrule
Llama2-7B & 61.79 & 63.85 & 51.49  & 40.09 & 8.23 \\[4pt]
Llama2-13B & 67.16 & 65.84 & 54.95 & 54.27 & 38.29 \\[4pt]
Llama2-70B & 76.39 & 84.62 &  66.42 & 55.29 & 56.21 \\[4pt]
Vicuna-v1.5 & 64.04 & 69.51 & 61.47  & 57.71 & 29.60 \\[4pt]
Vicuna-v1.5-13B & 68.96 & 80.24 & 71.90 & 68.23 & 42.14 \\[4pt]
WizardLM-13B & 67.49 & 76.33 & 55.62  & 54.14 & 42.11 \\[4pt]
InternLM-7B & 65.83 & 57.55 & 48.56  & 51.43 & 26.41 \\[4pt]
InternLM-20B & 85.20 & 86.33 & 83.79  & 81.41 & 82.91 \\[4pt]
Falcon-7b & 55.35 & 57.61 & 53.47 & 39.45 & 2.66 \\[4pt]
MPT-7B & 58.46 & 58.99 & 55.15 & 44.02 & 7.03 \\
\bottomrule
\end{tabular}
}
\label{table:boolq-ICL}
\end{center}
\end{table}

\begin{table}[ht]
\begin{center}
\caption{\revise{Comparisons of different attacks on the SocialIQA Dataset. In-context learning (ICL) improves the zero-shot performance, and the attack on in-context examples can decrease the performance. However, our adversarial attack has the biggest impact on the final performance (largest performance drop).}}
\resizebox{\linewidth}{!}{%
\begin{tabular}{lccccll}
\toprule
\textbf{Model} & \textbf{Original 0-shot} & \textbf{ICL} & \textbf{ICL Permutation} & \textbf{ICL Search} & \textbf{Permutation Attack} \\
\midrule
Llama2-7B & 52.00 & 57.63 & 46.37 & 33.52 & 15.71 \\[4pt]
Llama2-13B & 61.21 & 67.14 &  55.78 & 45.75 & 34.14 \\[4pt]
Llama2-70B & 71.60 & 75.64 & 66.99 & 64.53 & 49.85 \\[4pt]
Vicuna-v1.5 & 64.99 & 64.38 & 56.81  & 47.80 & 38.33 \\[4pt]
Vicuna-v1.5-13B & 66.07 & 68.58 &  58.96 & 51.07 & 44.42 \\[4pt]
WizardLM-13B & 63.46 & 62.64 &  50.97 & 43.19 & 31.78 \\[4pt]
InternLM-7B & 59.47 & 64.64 &  52.66 & 46.16 & 30.30 \\[4pt]
InternLM-20B & 79.48 & 78.86 & 75.49  & 70.47 & 65.97 \\[4pt]
Falcon-7b & 36.29 & 36.89 & 31.34 & 28.25 & 0.55 \\[4pt]
MPT-7B & 41.61 & 42.91 & 33.87 & 20.78 & 2.53 \\
\bottomrule
\end{tabular}
}
\label{table:socialiqa-ICL}
\end{center}
\end{table}

\begin{table}[ht]
\begin{center}
\caption{\revise{Comparisons of different attacks on the MedMCQA Dataset. In-context learning (ICL) improves the zero-shot performance, and the attack on in-context examples can decrease the performance. However, our adversarial attack has the biggest impact on the final performance (largest performance drop).}}
\resizebox{\linewidth}{!}{%
\begin{tabular}{lccccll}
\toprule
\textbf{Model} & \textbf{Original 0-shot} & \textbf{ICL} & \textbf{ICL Permutation} & \textbf{ICL Search} & \textbf{Permutation Attack} \\
\midrule
Llama2-7B & 37.96 & 39.67 & 32.10  & 26.57 & 1.60 \\[4pt]
Llama2-13B & 39.78 & 39.74 &  33.24 & 28.46 & 7.35 \\[4pt]
Llama2-70B & 49.61 & 51.78 &  41.37 & 41.76 & 7.35 \\[4pt]
Vicuna-v1.5 & 39.28 & 38.32 & 30.72  & 27.96 & 7.67 \\[4pt]
Vicuna-v1.5-13B & 41.80 & 43.22 & 36.72  & 30.68 & 11.90 \\[4pt]
WizardLM-13B & 34.87 & 36.54 &  26.24 & 23.59 & 6.32 \\[4pt]
InternLM-7B & 32.63 & 37.43 & 28.69  & 25.18 & 2.56 \\[4pt]
InternLM-20B & 43.61 & 42.58 &  38.14 & 33.20 & 13.92 \\[4pt]
Falcon-7b & 28.12 & 29.79 & 21.88 & 14.03 & 0.07 \\[4pt]
MPT-7B & 26.31 & 32.24 & 19.64 & 17.05 & 1.60 \\[4pt]
\bottomrule
\end{tabular}
}
\label{table:medmcqa-ICL}
\end{center}
\end{table}

\begin{table}[h]
\begin{center}
\caption{\revise{Results of using different sampling strategies and temperatures on MMLU dataset: before/after permutation.}}
\resizebox{\linewidth}{!}{%
\begin{tabular}{lcccccc}
\toprule
\textbf{Model} & \textbf{Greedy Decoding} &\textbf{ Temperature=0.5} & \textbf{Temperature=1.5} & \textbf{Top-k Sampling} & \textbf{Nucleus Sampling} \\
\midrule
Llama2-7B & 40.91/6.17 & 28.39/0.03 & 10.35/0.00 & 21.71/0.00 & 21.95/0.00 \\[4pt]
Llama2-13B & 52.22/18.33 & 44.00/3.67 & 13.94/0.00 & 32.54/0.00 & 32.42/0.02 \\[4pt]
Llama2-70B & 64.68/33.16 & 58.13/12.56 & 17.66/0.00 & 44.21/0.07 & 44.44/0.42 \\[4pt]
Vicuna-v1.5 & 48.57/18.09 & 47.64/12.29 & 34.43/0.04 & 42.71/3.60 & 44.77/8.10 \\[4pt]
Vicuna-v1.5-13B & 54.68/26.27 & 53.71/21.65 & 38.18/0.11 & 49.24/7.34 & 51.99/17.10 \\[4pt]
WizardLM-13B & 48.60/15.87 & 47.56/12.43 & 38.11/0.57 & 44.61/5.86 & 45.83/10.30 \\[4pt]
InternLM-7B & 45.72/10.45 & 0.01/0.00 & 0.53/0.00 & 0.16/0.00 & 0.07/0.00 \\[4pt]
InternLM-20B & 59.14/29.52 & 33.53/3.89 & 19.81/0.00 & 30.82/0.31 & 31.91/1.24 \\[4pt]
Falcon-7B & 31.66/2.49 & 0.02/0.00 & 0.46/0.00 & 0.06/0.00 & 0.01/0.00 \\[4pt]
MPT-7B & 35.60/3.52 & 0.01/0.00 & 0.67/0.00 & 0.12/0.00 & 0.04/0.00 \\
\bottomrule
\end{tabular}
}
\label{tab:temperatures}
\end{center}
\end{table}

\begin{table}[h]
\centering
\caption{\revise{Comparisons of position bias of Vicuna-13B with setting ground truth answers of in-context examples to specific positions.}}
\begin{tabular}{lcccc}
\toprule
 & A & B & C & D \\
\midrule
Original positional bias & 47.33 & \textbf{70.00} & 51.73 & 52.04 \\[4pt]
Moving ICL answers to A & 57.98 & \textbf{63.28} & 55.28 & 47.37 \\[4pt]
Moving ICL answers to B & 58.13 & \textbf{61.39} & 56.22 & 49.02 \\[4pt]
Moving ICL answers to C & 56.81 & \textbf{63.61} & 54.49 & 48.68 \\[4pt]
Moving ICL answers to D & 58.02 & \textbf{60.64} & 54.45 & 50.99 \\
\bottomrule
\end{tabular}
\label{tab:icl-vicuna13b}
\end{table}

\begin{table}[h]
\centering
\caption{\revise{Comparisons of position bias of InternLM-7B with setting ground truth answers of in-context examples to specific positions.}}
\begin{tabular}{lcccc}
\toprule
 & A & B & C & D \\
\midrule
Original positional bias & 45.72 & \textbf{37.23} & 65.12 & 41.49 \\[4pt]
Moving ICL answers to A & 31.55 & \textbf{74.76} & 44.70 & 43.42 \\[4pt]
Moving ICL answers to B & 38.71 & \textbf{73.49} & 43.94 & 42.47 \\[4pt]
Moving ICL answers to C & 31.58 & \textbf{69.05} & 49.27 & 46.47 \\[4pt]
Moving ICL answers to D & 32.63 & \textbf{69.39} & 44.34 & 51.22 \\

\bottomrule
\end{tabular}
\label{tab:icl-internlm7b}
\end{table}

\begin{table}[h]
\centering
\caption{\revise{Comparisons of position bias of InternLM-20B with setting ground truth answers of in-context examples to specific positions.}}
\begin{tabular}{lcccc}
\toprule
 & A & B & C & D \\
\midrule
Original positional bias & 51.05 & \textbf{68.75} & 53.47 & 62.35 \\[4pt]
Moving ICL answers to A & 51.33 & \textbf{72.52} & 62.81 & 55.29 \\[4pt]
Moving ICL answers to B & 49.70 & \textbf{73.07} & 64.34 & 56.31 \\[4pt]
Moving ICL answers to C & 48.01 & \textbf{70.00} & 64.36 & 60.61 \\[4pt]
Moving ICL answers to D & 46.99 & \textbf{67.54} & 62.20 & 65.72 \\
\bottomrule
\end{tabular}
\label{tab:icl-internlm20b}
\end{table}

\begin{table}[t]
\caption{\revise{Comparison of positional bias and our adversarial permutation attack on A-OKVQA dataset. While position bias exists, its impact is moderate. In contrast, our adversarial method severely degrades performance, usually below random chance level.}}
\begin{center}
\begin{tabular}{lclllll}
\toprule
{\bf Method}& {\bf Original} & {\bf A} & {\bf B} & {\bf C} & {\bf D} & {\bf Permutation Attack}\\
\midrule
  InstructBLIP-7B   &  74.06 &  67.16 & 75.28 & 75.90 & 75.11 & 51.62 \\[4pt]

  InstructBLIP-13B   &  77.90 & 77.29  & 72.75 & 80.35 & 73.54 & 55.38 \\[4pt]

  OpenFlamingo  &  64.68 &  52.34  & 72.77  &  41.19  & 35.86  &  3.58 \\[4pt]

 Otter-Llama7B  &  57.99 & 83.14 & 53.45 & 55.02  & 44.10 & 28.30 \\[4pt]

 Otter-MPT7B & 68.21  &  53.36 & 79.74  &  69.00 & 65.59  &  43.19  \\[4pt]
  
 LLaVA-7B & 52.91  & 77.18  & 22.71 & 14.85 & 10.94 & 0.09  \\[4pt]

 LLaVA-13B  & 63.14  &  69.43  & 77.79 & 63.76  & 48.08  & 25.85 \\[4pt]

 Limber  & 39.57  &  47.77  &  55.72 & 31.88 & 27.11  & 1.22 \\[4pt]

 mPLUG-Owl-pt &  39.91 &  33.26  & 45.16  &  47.57 &  36.49 & 1.83 \\[4pt]
   
 mPLUG-Owl-instr &  37.12 &  34.25  &  41.27 & 45.78   &  39.55 & 2.01 \\

\bottomrule
\end{tabular}
\end{center}
\label{table:VLLMs-pos}
\end{table}

\begin{table}[t]
\caption{\revise{Impact of majority vote, contextual calibration (C-Calibration), and maximum confidence (M-Confidence) defenses against the permutation attack on the A-OKVQA dataset. Contextual calibration fails completely. Majority vote and M-Confidence ameliorates the attack, but do not completely restore performance. Red shading indicates below-chance results.}}
\begin{center}
\resizebox{\linewidth}{!}{%
\begin{tabular}{lllllll}
\toprule
{\bf Method} & {\bf Original} & {\bf Adversarial Attack} & {\bf Majority Vote} & {\bf C-Calibration}  & {\bf \revise{M-Confidence}}  \\
\midrule

InstructBLIP-7B & 74.06 & 51.62 (\textcolor{red}{22.44 $\downarrow$}) & 57.47 (\textcolor{red}{16.59 $\downarrow$}) &  38.12 (\textcolor{red}{35.94 $\downarrow$}) &  69.79(\textcolor{red}{4.27 $\downarrow$})\\[4pt]

InstructBLIP-13B & 77.90 & 55.38 (\textcolor{red}{22.52 $\downarrow$}) & 60.26 (\textcolor{red}{17.64 $\downarrow$}) &  45.99 (\textcolor{red}{31.91 $\downarrow$}) & 70.83 (\textcolor{red}{7.07 $\downarrow$})\\[4pt]

OpenFlamingo & 46.90 &  \cellcolor{red!10}\phantom{0}3.58 (\textcolor{red}{43.32 $\downarrow$})  & \cellcolor{red!10}15.12 (\textcolor{red}{31.78 $\downarrow$}) & \cellcolor{red!10} 7.98 (\textcolor{red}{38.92 $\downarrow$}) & 44.20 (\textcolor{red}{2.70 $\downarrow$})\\[4pt]

Otter-Llama7B & 57.99 & 28.30 (\textcolor{red}{29.69 $\downarrow$}) &  27.63 (\textcolor{red}{30.36 $\downarrow$}) & \cellcolor{red!10}21.33 (\textcolor{red}{36.66 $\downarrow$}) & 38.29 (\textcolor{red}{19.70 $\downarrow$})\\[4pt]

Otter-MPT7B & 68.21 & 43.19 (\textcolor{red}{25.02 $\downarrow$}) & 55.11  (\textcolor{red}{13.10 $\downarrow$}) & 42.46 (\textcolor{red}{25.75 $\downarrow$}) & 51.97 (\textcolor{red}{16.24 $\downarrow$})\\[4pt]

LLaVA-7B & 52.91 &\cellcolor{red!10}\phantom{0}0.09 (\textcolor{red}{52.82 $\downarrow$}) & 27.86 (\textcolor{red}{25.05 $\downarrow$}) & \cellcolor{red!10}\phantom{0}8.23 (\textcolor{red}{44.68 $\downarrow$}) & 50.04 (\textcolor{red}{2.87 $\downarrow$}) \\[4pt]

LLaVA-13B & 63.14 & 29.52 (\textcolor{red}{33.62 $\downarrow$}) & 53.36 (\textcolor{red}{9.78 $\downarrow$}) & 28.94 (\textcolor{red}{34.20 $\downarrow$}) & 64.80 (\textcolor{blue}{1.66 $\uparrow$})\\[4pt]

Limber & 39.57 & \cellcolor{red!10}\phantom{0}1.22 (\textcolor{red}{38.35 $\downarrow$}) & 38.69 (\textcolor{red}{0.88 $\downarrow$}) & \cellcolor{red!10}\phantom{0}3.59 (\textcolor{red}{35.98 $\downarrow$}) & 38.14 (\textcolor{red}{1.43 $\downarrow$}) \\[4pt]

mPLUG-Owl-pt & 39.91 & \cellcolor{red!10}\phantom{0}1.83 (\textcolor{red}{38.08 $\downarrow$}) & \cellcolor{red!10}14.33 (\textcolor{red}{25.58 $\downarrow$}) & \cellcolor{red!10}\phantom{0}4.28 (\textcolor{red}{35.63 $\downarrow$}) & \cellcolor{red!10}15.21 (\textcolor{red}{24.70 $\downarrow$})\\[4pt]

mPLUG-Owl-instr & 37.12 & \cellcolor{red!10}\phantom{0}2.01 (\textcolor{red}{35.11 $\downarrow$}) & \cellcolor{red!10}12.01 (\textcolor{red}{25.11 $\downarrow$}) & \cellcolor{red!10}\phantom{0}2.19 (\textcolor{red}{34.93 $\downarrow$}) & \cellcolor{red!10}13.37 (\textcolor{red}{23.75 $\downarrow$})\\

\bottomrule
\end{tabular}
}
\end{center}
\label{table:maj-cbu-vllm}
\end{table}

\begin{table*}[h]
\caption{Qualitative results of permutations of answer options and corresponding predictions (Llama2-7B) from ARC-challenge dataset.}
\vspace{-2pt}
\begin{center}
\resizebox{0.9\linewidth}{!}{%
\begin{tabular}{l}
\toprule
\textbf{Question:} A physicist wants to determine the speed a car must reach to jump over a ramp. The physicist conducts three trials.\\In trials two and three,
the speed of the car is increased by 20 miles per hour. What is the physicist investigating when he changes the speed? \\[2pt]
\textbf{True Answer:} the independent (manipulated) variable.\\[2pt]
\midrule
\textbf{Original Answer Set:} A. the control B. the hypothesis statement C. the dependent (responding) variable \underline{D. the independent (manipulated) variable.} \\[2pt]
\hspace{1cm} Model Prediction: \textcolor{darkgreen}{D}. \\[2pt]
\textbf{Permutation 1:} A. the control B. the hypothesis statement \underline{C. the independent (manipulated) variable} D. the dependent (responding) variable  \\[2pt]
\hspace{1cm} Model Prediction: \textcolor{darkgreen}{C}. \\[2pt]
\textbf{Permutation 2:} A. the hypothesis statement B. the control C. the dependent (responding) variable \underline{D. the independent (manipulated) variable.} \\[2pt]
\hspace{1cm} Model Prediction: \textcolor{red}{A}. \\[2pt]
\bottomrule
\end{tabular}
}
\end{center}
\label{table:lqa}
\end{table*}

\end{document}